\documentclass[sigplan,10pt]{acmart}
\acmSubmissionID{351}
\copyrightyear{2027}
\acmYear{2027}
\setcopyright{cc}
\setcctype{by}
\acmConference[EuroSys '27]{22nd European Conference on Computer Systems}{April 19--23, 2027}{Rabat, Morocco}
\acmBooktitle{22nd European Conference on Computer Systems (EuroSys '27), April 19--23, 2027, Rabat, Morocco}
\acmDOI{10.1145/3842654.3848543}
\acmISBN{979-8-4007-2971-3/2027/04}
\acmBadgeR[https://www.acm.org/publications/policies/artifact-review-and-badging-current]{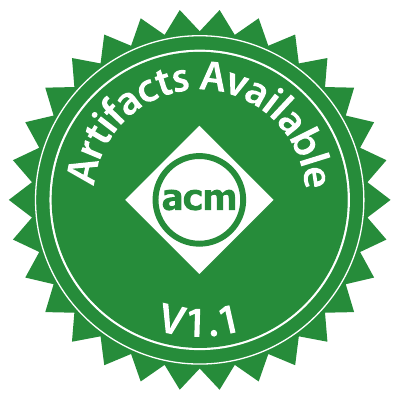}
\acmBadgeR[https://www.acm.org/publications/policies/artifact-review-and-badging-current]{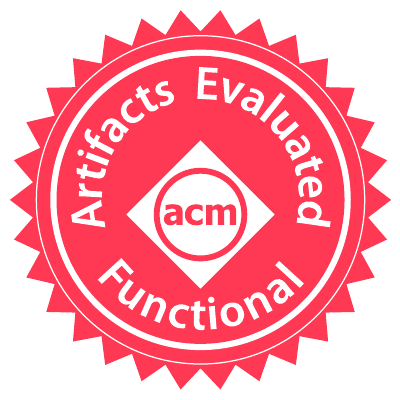}


\usepackage{tikz}
\usepackage{amsmath}
\usepackage{array}

\usepackage{filecontents}

\usepackage{subfiles}
\usepackage{layouts}
\usepackage{multirow}
\usepackage{subcaption}
\usepackage{wrapfig}
\usepackage{amsfonts}
\usepackage{makecell}
\usepackage{algorithm}
\usepackage[noend]{algpseudocode}
\usepackage{booktabs}
\usepackage{enumitem}
\usepackage{listings}

\newcommand{\revadd}[1]{{\leavevmode\color{black}#1}}
\definecolor{revproteuscolor}{HTML}{000000}
\newcommand{\revproteus}[1]{{\leavevmode\color{revproteuscolor}#1}}
\definecolor{revshepcolor}{HTML}{000000}
\newcommand{\revshep}[1]{{\leavevmode\color{revshepcolor}#1}}
\newcommand{\revrm}[1]{{\color{blue}\sout{}}}
\usepackage{ulem}
\usepackage{float}

\newcolumntype{P}[1]{>{\centering\arraybackslash}p{#1}}
\algrenewcommand\algorithmiccomment[1]{\hspace{1em}// #1}
\algrenewcommand\algorithmicindent{1.0em}
\begin{document}

\title{TeDiServe: High SLO Attainment Serving for Diffusion Language Models}

\author{Tzu-Tao Chang}
\affiliation{%
  \institution{University of Wisconsin-Madison}
  \city{Madison}
  \state{WI}
  \country{USA}
}
\email{tchang85@wisc.edu}

\author{Benjamin Yuanyang Hong}
\affiliation{%
  \institution{University of Wisconsin-Madison}
  \city{Madison}
  \state{WI}
  \country{USA}
}
\email{byhong@wisc.edu}

\author{Kiet Pham}
\affiliation{%
  \institution{University of Wisconsin-Madison}
  \city{Madison}
  \state{WI}
  \country{USA}
}
\email{kvpham@wisc.edu}

\author{Shivaram Venkataraman}
\affiliation{%
  \institution{University of Wisconsin-Madison, ETH Zurich}
  \city{Zurich}
  \country{Switzerland}
}
\email{shivaram@cs.wisc.edu}

\begin{abstract}

Diffusion language models (DLMs) have recently emerged as a promising alternative to conventional autoregressive language models. By generating multiple tokens in parallel during each denoising step, they offer higher inference throughput while maintaining competitive quality. However, realizing these throughput gains while meeting latency SLOs in a serving system requires addressing challenges introduced by DLMs' unique characteristics. These include navigating the speed-quality tradeoff created by confidence-based denoising, choosing appropriate parallelization levels across model instances under fluctuating load, and coordinating approximate KV caching mechanisms that introduce non-uniform per-step costs. To address these challenges, we present TeDiServe, a cluster-level serving system for DLMs. TeDiServe enables deadline-aware scheduling and adaptive load control through confidence-threshold adjustment, and dynamically reconfigures the cluster by solving a quality-aware optimization problem, while explicitly modeling the step-level heterogeneity introduced by approximate KV caching. Across multiple benchmarks and real-world traces, TeDiServe improves SLO attainment by up to 56.6 percentage points and reduces end-to-end request latency by up to 46\% while incurring less than 1\% accuracy drop.

\end{abstract}

\maketitle

\section{Introduction}
\label{sec:intro}
Large language models (LLMs) exhibit strong general-purpose capabilities and are increasingly integrated into  services across domains, such as conversational agents~\cite{openai2024gpt4}, code generation~\cite{anthropic_claude_code}, retrieval-augmented generation~\cite{gao2023retrieval}, and agentic task execution~\cite{zheng2025deepresearcher}. As adoption scales, efficiently serving LLM workloads becomes increasingly important and challenging. Service providers must sustain high throughput to handle massive request volumes (billions of queries per day~\cite{roth2025chatgptstats}) while simultaneously meeting strict latency SLOs to maintain interactive responsiveness.

\begin{figure}[t]
    \centering
    \includegraphics[width=0.85\linewidth]{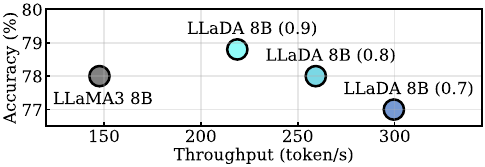}
    \vspace{-1ex}
    \caption{Accuracy and throughput comparison of LLaMA~\cite{touvron2023llama} (autoregressive) and LLaDA~\cite{nie2025llada} (diffusion) on GSM8K~\cite{cobbe2021gsm8k}, run on a single H100 GPU. Numbers in parentheses denote the denoising confidence thresholds.}
    \label{fig:intro/ar_vs_dllm}
    \vspace{-8pt}
\end{figure}

While many system-level optimizations have been proposed to improve the efficiency of LLM serving~\cite{zhong2024distserve, kwon2023vllm, stojkovic2025dynamollm, sun2024llumnix, zhu2025nanoflow}, the throughput of current deployments is fundamentally constrained by the \emph{autoregressive} nature of these models: tokens must be generated sequentially, one at a time. This inherent serialization limits parallelism and makes it difficult to scale throughput to meet growing request loads. In response to these limitations, researchers and industry have explored \emph{diffusion language models} (DLMs) as a powerful and promising alternative~\cite{nie2025llada, ye2025dream, google2025geminidiffusion, inceptionlabs2025mercury, song2025seeddiffusion}. These models generate text through an iterative refinement process: starting from a fully corrupted sequence containing only \texttt{[MASK]} tokens, the model repeatedly performs denoising steps, unmasking all tokens with confidence exceeding a certain threshold at each step. Because multiple tokens can be generated in parallel during each iteration, DLMs achieve higher throughput than classic autoregressive language models while maintaining comparable generation quality. Prior industry reports~\cite{inceptionlabs2025mercury} have shown that DLMs can achieve 10$\times$ greater throughput and our own measurements (Figure~\ref{fig:intro/ar_vs_dllm}) show that \revadd{open-weight} DLMs achieve up to \revadd{1.75$\times$} higher throughput at equal or better accuracy compared to similarly sized autoregressive models.

\begin{figure}[t]
    \centering
    \includegraphics[width=0.85\linewidth]{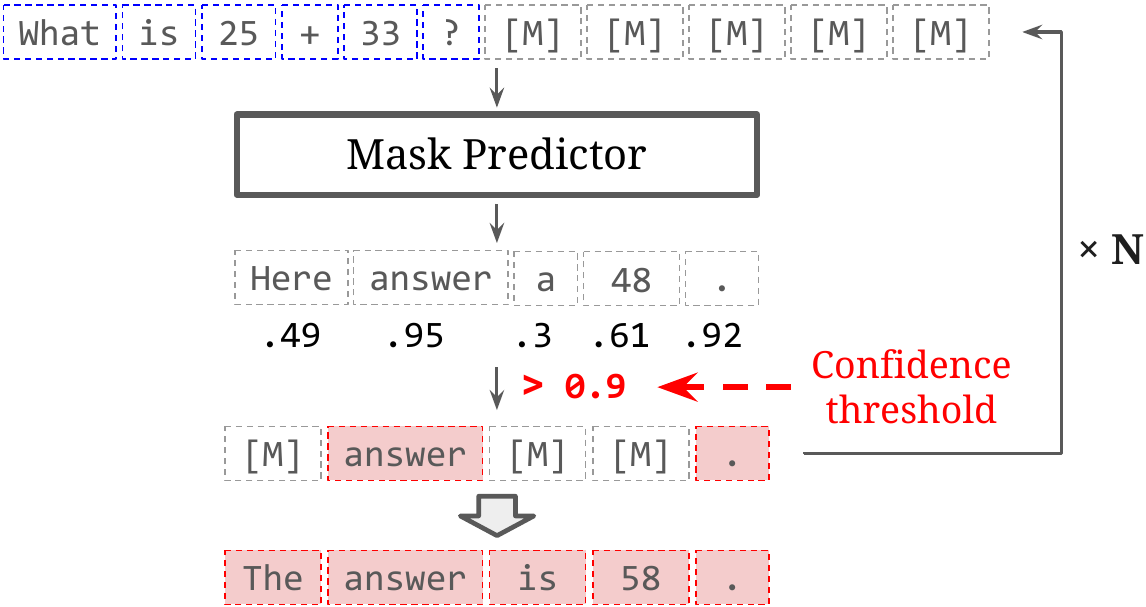}
    \vspace{-3pt}
    \caption{Diffusion language models.}
    \label{fig:dllm}
    \vspace{-8pt}
\end{figure}

However, realizing DLM throughput gains while meeting latency SLOs requires addressing several challenges. First, the use of token confidence in the denoising process (Figure~\ref{fig:dllm}) introduces an explicit speed-quality knob: lowering the confidence threshold can reduce the number of denoising steps required, at the cost of a modest quality drop. This means that DLM serving systems should choose the appropriate threshold for each request to meet SLOs while ensuring response quality. Second, each denoising step in DLMs is compute-heavy~\cite{wu2025fastdllm, wu2025fastdllmv2} making end-to-end latency and throughput sensitive to the degree of parallelization used; with a fixed number of GPUs, using a higher parallelization degree lowers per-request latency but leads to lower aggregate throughput (Figure~\ref{fig:tp_motivation}). This means that DLM serving systems need to jointly select the confidence threshold and parallelization degree, taking into account fluctuating workload conditions.

To address these challenges, we present TeDiServe \revadd{(\textbf{Te}xt \textbf{Di}ffusion \textbf{Serv}ing)}, a cluster-level serving system for DLMs that achieves high throughput while meeting strict latency SLOs. TeDiServe manages a cluster of model instances and makes scheduling decisions at the granularity of denoising steps, enabling fine-grained load balancing and deadline-aware adaptation. For each request, TeDiServe dynamically adjusts the confidence threshold to balance throughput, latency, and generation quality. It uses SLO-aware threshold selection to identify the highest confidence threshold predicted to meet the request’s remaining latency budget, thereby maximizing quality subject to the deadline. In addition, it applies adaptive load control to bound the maximum threshold each request may use so that the aggregate workload remains within the cluster’s processing capacity. Given the resulting threshold, TeDiServe schedules the next denoising step on the instance with the lowest step completion time, accounting for per-instance load and tensor parallelism (TP) degree to minimize execution latency.

To adapt the cluster configuration to changing workloads, TeDiServe periodically solves a two-stage integer linear program (ILP). In the first stage, it determines the number of model instances at each TP degree to ensure sufficient steady-state capacity while maximizing average confidence threshold and minimizing latency. If the resulting configuration differs from the current one, the second stage computes a reconfiguration plan that minimizes migration overhead, thereby minimizing disruption to service availability. In addition, prior work has proposed approximate key-value (KV) caching for DLMs, which reuses KV states across denoising steps to reduce redundant computation but requires regular refreshes to avoid staleness. To account for the non-uniform computational cost introduced by approximate KV caching, TeDiServe incorporates this step-level heterogeneity directly into its scheduling and reconfiguration decisions.

We implement TeDiServe on top of vLLM~\cite{kwon2023vllm} and Ray~\cite{moritz2018ray}, extending them to support DLMs, multi-instance serving, and approximate KV caching. \revadd{Evaluation on a real-world LLM serving trace shows that TeDiServe improves SLO attainment by 30.2 percentage points and reduces latency by 46\% with only a 0.09 drop in quality score. Across multiple accuracy benchmarks, it improves SLO attainment by up to 56.6 percentage points with only a 0.9\% accuracy drop under high load.}

\revshep{\textbf{Our contributions.} We make the following key contributions:
\begin{itemize}[leftmargin=1.2em, itemsep=0.1em, topsep=0.2em]
    \item We identify that the confidence threshold of a DLM is not merely a decoding hyperparameter, but a fine-grained runtime control knob for meeting latency SLOs, which exposes a serving opportunity specific to this emerging model family.
    \item We design a scheduler with two-level confidence-threshold control that jointly considers each request's remaining latency budget and system-wide load, while accounting for heterogeneous step times introduced by approximate KV caching.
    \item We show that per-token confidence values are strong indicators of remaining denoising steps and use them to build a lightweight predictor for deadline-aware threshold selection.
    \item We design a quality-aware two-stage MILP that reconfigures the cluster to the TP configuration best suited to the current workload.
    \item We implement TeDiServe and show that it improves SLO attainment by up to 56.6 percentage points over serving systems adapted for DLMs, with at most a 0.9\% quality reduction.
\end{itemize}}
\vspace{-1ex}
\section{Background and Challenges}

\subsection{Diffusion Language Models}
Diffusion has emerged as a promising alternative paradigm to autoregressive language modeling~\cite{sahoo2024mdlm, austin2021d3pm, lou2024sedd}.
Rather than generating tokens one-by-one in a left-to-right order, DLMs initialize generation from a corrupted sequence and progressively refine it through a sequence of denoising steps~\cite{austin2021d3pm, sahoo2024mdlm, he2023diffusionbert}, analogous to diffusion image models. More specifically, at each denoising step, the prompt, concatenated with a user-specified number of \texttt{[MASK]} tokens, is fed into the model’s mask predictor. The mask predictor produces predictions for all masked tokens in parallel in a single forward pass, after which the model selectively keeps a subset of tokens and remasks the rest. The partially unmasked sequence is then used as input to the next denoising step, and this iterative refinement continues until no \texttt{[MASK]} tokens remain.

Existing DLMs predominantly use Transformer-based mask predictors with full bidirectional attention~\cite{nie2025llada, ye2025dream, inceptionlabs2025mercury}. This design allows the predictor to condition on both left and right context when inferring each masked token, and has been shown to provide stronger capabilities on tasks involving reversal reasoning~\cite{berglund2024reversal}.

A common and effective strategy for deciding which tokens to unmask is based on token confidence~\cite{nie2025llada,wu2025fastdllm,cheng2025sdar}. For each predicted token, the model uses its softmax probability as a confidence score, and at each denoising step unmasks only those tokens whose confidence exceeds a predefined threshold. If no predicted tokens have confidence exceeding the threshold, the token with the highest confidence is unmasked to ensure progress. Figure~\ref{fig:dllm} illustrates this process. 

This confidence-based refinement progressively resolves uncertain regions and often produces more coherent text~\cite{li2025surveydlm}. Moreover, because multiple tokens may exceed the confidence threshold at each step, diffusion models can unmask several tokens in parallel, achieving higher throughput than autoregressive models while maintaining comparable quality. Figure~\ref{fig:intro/ar_vs_dllm} shows that DLMs achieve 1.75$\times$ higher throughput at equal or better accuracy compared to similarly sized autoregressive models. These advantages make DLMs a compelling alternative to traditional autoregressive approaches and have driven significant recent progress, including both large open-source models~\cite{nie2025llada, ye2025dream, zhu2025llada15, zhu2025lladamoe} and production-scale systems~\cite{inceptionlabs2025mercury, song2025seeddiffusion, google2025geminidiffusion}.
\vspace{-6pt}
\subsection{Challenges and Opportunities for Efficient Diffusion Language Model Serving}
While DLMs offer promising capabilities, their unique characteristics also introduce new challenges and opportunities for efficient serving. To fully leverage these models in practice, we outline three key challenges and opportunities that a DLM serving system must address.

\subsubsection{Speed-Quality Trade-off}
\label{subsubsec:challenges/speed_quality_tradeoff}
\begin{figure}[t]
    \centering
    \begin{subfigure}[t]{0.46\linewidth}
        \centering
        \includegraphics[width=\linewidth]{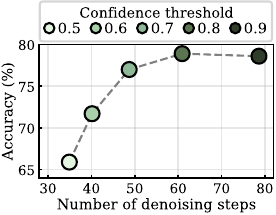}
        \caption{Fixed confidence threshold: each dot corresponds to using a single threshold throughout the entire denoising process.}
        \label{fig:accuracy_vs_step_fixed}
    \end{subfigure}
    \hfill
    \begin{subfigure}[t]{0.46\linewidth}
        \centering
        \includegraphics[width=\linewidth]{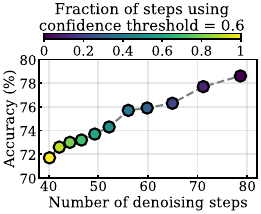}
        \caption{Dynamic confidence threshold: the default threshold is 0.9; each point uses a randomly chosen fraction of steps that switch to 0.6.}
        \label{fig:accuracy_vs_step_dynamic}
    \end{subfigure}
    \vspace{-1ex}
    \caption{\small{Accuracy-speed tradeoff of LLaDA~\cite{nie2025llada} on GSM8K~\cite{cobbe2021gsm8k}}}
    \label{fig:accuracy_vs_step_tradeoff}
    \vspace{-8pt}
\end{figure}
The iterative denoising process in DLMs introduces an inherent speed-quality trade-off. At each denoising step, only tokens whose confidence exceeds a predefined confidence threshold are unmasked. With a high threshold, the model only unmasks tokens it is very confident in, which improves correctness but reduces the number of tokens unmasked per step, increasing the total number of denoising steps. With a low threshold, more tokens can be unmasked per step, reducing the number of steps needed but increasing the risk of unmasking incorrect tokens and hurting quality. This is illustrated in Figure~\ref{fig:dllm}: if the threshold is set to 0.6, the token ``48'' would be unmasked for the question ``What is 25 + 33?'', which is incorrect. Figure~\ref{fig:accuracy_vs_step_fixed} shows this trade-off for LLaDA~\cite{nie2025llada} on the GSM8K~\cite{cobbe2021gsm8k} benchmark. Using a lower threshold reduces both accuracy and the number of steps. 

In addition, the confidence threshold does not need to remain fixed throughout the denoising process; the speed-quality trade-off can be controlled at the granularity of individual denoising steps. Figure~\ref{fig:accuracy_vs_step_dynamic} illustrates this effect by showing accuracy and total denoising steps when mixing a high confidence threshold (0.9) with a lower threshold (0.6). As the fraction of steps using the lower threshold increases, both the achieved accuracy and the number of denoising steps decrease accordingly.

\revadd{This trade-off creates opportunities for improved serving performance, both for meeting individual request SLOs and for system-wide load control. At the request level, this knob enables deadline-aware adaptation: if a request is predicted to miss its SLO under the current threshold, the server can temporarily lower the threshold to reduce the remaining number of denoising steps, then return to a higher threshold when slack permits. At the system level, a load-aware controller can lower thresholds during overload to reduce per-request work and keep the aggregate load within what the cluster can sustainably handle. This stabilizes queue growth and prevents the tail-latency cascades that otherwise lead to broad SLO misses. As shown in Figure~\ref{fig:accuracy_vs_step_fixed}, reducing the threshold from 0.9 to 0.7 decreases the number of steps by 38\%, suggesting that the system can process 1.61$\times$ more requests with only a 1.5\% accuracy drop. In both cases, the quality impact remains bounded as long as the thresholds stay within an acceptable range, allowing us to trade small, controlled reductions in quality for meaningful latency improvements.}
\subsubsection{Cluster-Level Serving}
\label{subsubsec:challenges/cluster_level_serving}
\begin{figure}[t]
    \centering
    \begin{subfigure}[t]{0.46\linewidth}
        \centering
        \includegraphics[width=\linewidth]{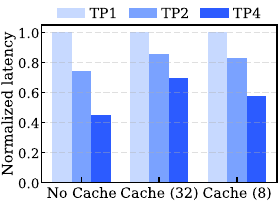}
        \caption{\revadd{Latency under different TP degrees. Numbers in parentheses indicate the denoising block size.}}
        \label{fig:tp_effect}
    \end{subfigure}
    \hfill
    \begin{subfigure}[t]{0.46\linewidth}
        \centering
        \includegraphics[width=\linewidth]{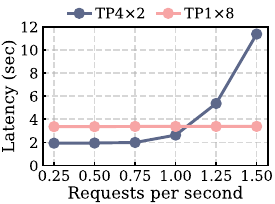}
        \caption{Latency under 2 TP-4 instances vs.\ 8 TP-1 instances across RPS.}
        \label{fig:config_comparison}
    \end{subfigure}
    \vspace{-2ex}
    \caption{Tensor parallelism for DLMs.}
    \label{fig:tp_motivation}
    \vspace{-3ex}
\end{figure}
With full bidirectional attention, each denoising step in a DLM is computationally intensive. As a result, these models benefit substantially from tensor parallelism. Figure~\ref{fig:tp_effect} shows that employing tensor parallelism reduces per-step latency by \revadd{25\%} with 2 GPUs and by \revadd{55\%} with 4 GPUs. In a cluster-serving setting, this suggests that high-TP-degree workers can significantly reduce end-to-end latency; migrating requests to these workers can therefore improve the likelihood of meeting SLOs when requests are close to violating their deadlines.

However, with a fixed cluster size, choosing the TP degree for workers introduces a fundamental latency-throughput trade-off. Figure~\ref{fig:config_comparison} illustrates two cluster configurations: one with fewer high-TP-degree workers and one with more low-TP-degree workers. Under low load, the high-TP configuration reduces request latency by up to 43\% by leveraging model parallelism. Under high load, however, its limited number of workers constrains overall throughput, causing the request queue to grow rapidly. \revshep{Concretely, in this 8-GPU example, a TP-4 instance serves a request in 2~s and a TP-1 instance in 3.6~s, giving throughputs of 0.5 and 0.28 requests per second; the two TP-4 instances therefore sustain only 1.0 requests per second in aggregate, compared to 2.22 for the eight TP-1 instances.} Once the arrival rate exceeds this capacity, the resulting queueing delay dominates end-to-end latency, leading to substantial slowdown. In contrast, the low-TP configuration remains stable due to its higher throughput capacity. Real-world serving traces show that request arrival rates and compositions fluctuate significantly over time~\cite{wang2025burstgpt, hu2025deepserve}, implying that a single, static cluster configuration is insufficient.

The ability to adjust confidence thresholds adds another dimension to this problem. When load increases, the system has two levers: it can lower confidence thresholds to reduce per-request computation at the cost of potential quality degradation, or it can reconfigure the cluster toward lower TP degrees to increase throughput at the cost of higher per-step latency. These interacting trade-offs between latency, throughput, and quality make static policies insufficient. Serving DLMs effectively therefore requires dynamic reconfiguration strategies that jointly account for threshold control, TP-degree allocation, and workload conditions.
\subsubsection{Support for Approximate Caching}
One widely used optimization in Transformer-based autoregressive models is KV caching. With autoregressive models, each newly generated token attends only to previously generated tokens; as a result, the key and value tensors for the prefix remain invariant throughout decoding. These tensors can therefore be cached and reused exactly across decoding steps, eliminating redundant computation and reducing inference cost. In contrast, DLMs cannot leverage exact KV caching. Because they employ bidirectional attention, the key and value tensors depend on the entire (partially masked) sequence at each denoising step and thus change as the sequence is progressively refined. This makes the standard KV-cache optimization used in autoregressive models inapplicable. 

To address this limitation, recent work~\cite{wu2025fastdllm, liu2025dllmcache} introduces approximate KV-caching mechanisms tailored for DLMs. These approaches partition the output sequence into blocks, whose size is configured beforehand, and process them sequentially. Within each block, the model runs multiple denoising steps to gradually unmask tokens. A key empirical observation is that, during the denoising of a given block, the key and value activations for the prompt, previously completed blocks (prefix blocks), and future blocks that remain fully masked (suffix blocks) change only minimally across consecutive steps. Leveraging this stability, these methods cache and reuse the KV tensors for the prefix and suffix blocks throughout the denoising steps of the current block. To avoid excessive staleness, the KV tensors for all blocks are recomputed after the current block is fully unmasked, before the model processes the next block.

By avoiding redundant computation, approximate KV caching can reduce latency by up to 2.1$\times$ without significant degradation in generation quality~\cite{wu2025fastdllm}. A practical serving system for DLMs must therefore support approximate KV caching while also coordinating it with dynamic confidence-threshold control and cluster reconfiguration.

\section{TeDiServe}
\label{sec:tediserve}
TeDiServe is a cluster-level serving system for DLMs that achieves high SLO attainment by dynamically adjusting the denoising-step confidence threshold and using an ILP-based reconfiguration framework tailored to the characteristics of DLMs. In this section, we first provide an overview of TeDiServe's design and architecture in \S\ref{subsec:tediserve/overview}. We then describe the scheduling and confidence-adjustment algorithm in \S\ref{subsec:tediserve/scheduling}. \S\ref{subsec:tediserve/reconfiguration} describes how TeDiServe dynamically reconfigures the cluster in response to workload changes. \S\ref{subsec:tediserve/kv_caching} explains how these ideas generalize to support approximate KV caching. We conclude by discussing extensions for more complex SLOs and priority policies in \S\ref{subsec:tediserve/extension}.
\vspace{-4pt}
\subsection{Design Overview}
\label{subsec:tediserve/overview}
\begin{figure}[t]
    \centering
    \includegraphics[width=0.98\linewidth]{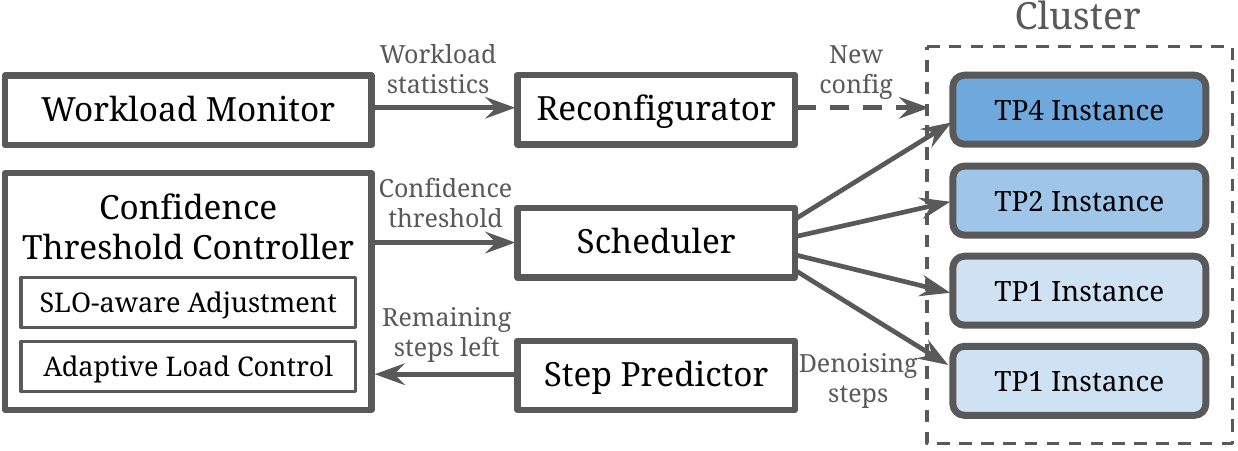}
    \vspace{-3pt}
    \caption{TeDiServe architecture.}
    \label{fig:architecture}
    \vspace{-8pt}
\end{figure}
Figure~\ref{fig:architecture} shows the architecture of TeDiServe. The system administrator specifies the request latency SLO (we assume a homogeneous SLO for simplicity but discuss extensions in \S\ref{subsec:tediserve/extension}), the set of supported TP degrees, and the initial cluster configuration (i.e., the number of model instances and their TP degrees). The administrator also specifies a set of candidate confidence thresholds for token unmasking \revadd{(e.g., [0.5, 0.6, 0.7, 0.8, 0.9])}; this set should be chosen per model and application to provide sufficient control flexibility while ensuring acceptable quality even at the lowest threshold. The system performs a one-time latency profiling step to measure the per-denoising-step latency for each batch size under every supported TP degree, similar to existing systems~\cite{stojkovic2025dynamollm, sun2024llumnix}. These measurements guide TeDiServe’s scheduling and reconfiguration decisions.

\revadd{For each request, the Confidence Threshold Controller determines the confidence threshold used for future denoising steps to maximize generation quality while ensuring the request meets its latency SLO. To achieve this, it selects the highest threshold from the candidate set that keeps the request within its remaining SLO budget. The threshold is adjusted dynamically throughout the denoising process. However, making these decisions independently for each request is insufficient: if all requests select the highest feasible threshold without accounting for one another, the resulting aggregate load can exceed cluster capacity, causing queue growth and system-wide SLO violations. The Confidence Threshold Controller therefore also applies adaptive load control that limits the maximum threshold each request may use so that the aggregate load remains within cluster capacity while preserving as much flexibility for high-quality generation as possible.}

\revadd{The Confidence Threshold Controller relies on estimates of the remaining denoising steps for each request. These estimates are produced by the Step Predictor, a lightweight model trained offline to predict the remaining denoising steps based on the request's current state, including its unmasking progress and token-confidence statistics.}

\revadd{Once the confidence threshold is determined for a request, the Scheduler assigns each denoising step to the model instance that yields the lowest step latency (e.g., higher TP degree or lower load), and migrates requests when doing so reduces latency. Each model instance performs denoising asynchronously for its assigned batch of requests using the confidence threshold selected by the Confidence Threshold Controller. After each step, the newly unmasked tokens and confidence statistics are returned to and tracked by the Scheduler, which then makes scheduling decisions for the next denoising step.}

The Workload Monitor tracks request arrivals and their input and output length statistics. The Reconfigurator uses this information to periodically compute a cluster configuration suited to the current workload. This is formulated as a two-stage ILP. In the first stage, the objective is to maximize the per-token confidence threshold (thereby maximizing generation quality) while ensuring steady-state capacity sufficiency; the output of this stage is the desired number of model instances at each TP degree. If the current cluster configuration does not match this plan, the second stage determines how to transition to the new configuration with minimal overhead by identifying which existing model instances to terminate and on which nodes to launch new ones.
\subsection{Dynamic Confidence Threshold Adjustment and Scheduling}
\label{subsec:tediserve/scheduling}
\revadd{This section describes the core confidence-threshold adjustment and scheduling mechanisms in TeDiServe. We first describe how the Confidence Threshold Controller selects the confidence threshold that satisfies each request's remaining SLO budget (\S\ref{subsubsec:tediserve/threshold_adjustment/slo_aware_conf_adjustment}). We then present the adaptive load control mechanism that limits the maximum confidence threshold used for each request to keep the aggregate load within cluster capacity (\S\ref{subsubsec:tediserve/threshold_adjustment/load_control}). Next, we describe the Step Predictor used to estimate the remaining denoising steps for each request (\S\ref{subsubsec:tediserve/threshold_adjustment/step_prediction}). Finally, we present the scheduling algorithm that assigns denoising steps to model instances based on TP degree and current load (\S\ref{subsubsec:tediserve/threshold_adjustment/scheduling}).}
\subsubsection{\texorpdfstring{\revadd{SLO-aware Confidence Threshold Adjustment}}{SLO-aware Confidence Threshold Adjustment}}
\label{subsubsec:tediserve/threshold_adjustment/slo_aware_conf_adjustment}
\begin{algorithm}[t]
\caption{SLO-aware Confidence Threshold Adjustment}
\label{alg:selecting_conf_thresh}
\begin{algorithmic}[1]

\Function{PredictedTimeLeft}{req, $\gamma$}
    \State steps $\gets$ req.predStepsLeft[$\gamma$]
    \State tp    $\gets$ req.instance.tpDegree
    \State batch $\gets$ req.instance.numTokens
    \State perStepLatency   $\gets$ latencyProfile[tp][batch]
    \State \Return steps $\times$ perStepLatency
\EndFunction

\Function{UpdateConfThreshold}{req}
    \For{$\gamma$ in desc req.allowedConfThresh}
        \State req.confThreshold $\gets$ $\gamma$
        \If{\Call{PredictedTimeLeft}{req, $\gamma$} $\le$ req.sloLeft}
            \State \textbf{break}
        \EndIf
    \EndFor
\EndFunction
\end{algorithmic}
\end{algorithm}
\revadd{Given a candidate set of confidence thresholds, our goal is to maximize generation quality while ensuring a request does not violate its latency SLO. As shown in Figure~\ref{fig:accuracy_vs_step_dynamic}, generation quality improves when more denoising steps use a high confidence threshold. This motivates selecting the highest allowable threshold whenever doing so does not violate the request's SLO.}

\revadd{To determine whether an SLO would be violated, we estimate the remaining latency of a request based on (1) the predicted number of denoising steps left under a given confidence threshold, obtained from the Step Predictor, and (2) the per-step iteration time of the assigned model instance, obtained from the latency profile for its TP degree and batch size. Using this estimation, the system selects the highest confidence threshold within the allowed range whose estimated remaining latency does not exceed the request's remaining SLO budget. This procedure is outlined in Algorithm~\ref{alg:selecting_conf_thresh}. As shown in Section~\ref{subsec:eval/ablation}, this SLO-aware confidence threshold adjustment improves SLO attainment by up to 70.8 percentage points under high load compared to using a fixed confidence threshold.}
\subsubsection{\texorpdfstring{\revadd{Adaptive Load Control}}{Adaptive Load Control}}
\label{subsubsec:tediserve/threshold_adjustment/load_control}
\revadd{While SLO-aware threshold adjustment improves each request individually, it lacks awareness of overall system conditions. If every request independently chooses the highest threshold that satisfies its own SLO, the resulting aggregate load can exceed the cluster's processing capacity, causing long queueing delays and widespread SLO violations. As shown in Section~\ref{subsec:eval/ablation}, this could lead to a reduction in SLO attainment by up to 42.4 percentage points. Adaptive load control prevents this by restricting the maximum confidence threshold each request may use so that the aggregate load remains within the total processing capacity of the cluster over the SLO window.

More specifically, the work needed for a request is the product of its total number of tokens (input plus output) and its predicted number of remaining denoising steps, and the aggregate load is the sum of this work across all active requests. A model instance's capacity over the SLO window is the number of tokens it can process per second multiplied by the SLO duration, and the total cluster capacity is the sum of the capacities of all model instances.}

\revadd{Given a fixed cluster configuration, the total capacity is constant, whereas the aggregate load depends on the confidence thresholds used. Our goal is thus to give each request the widest feasible confidence-threshold range by assigning it the highest allowable maximum threshold, so that the per-request controller can still choose high thresholds for better generation quality, while ensuring that the worst-case aggregate load -- where every request uses its maximum permitted threshold -- remains within the cluster's capacity.}

\begin{algorithm}[!t]
\caption{Adaptive Load Control}
\label{alg:update_allowed_conf_thresh}
\begin{algorithmic}[1]

\Function{ReqLoad}{req, $\gamma$}
    \State \Return req.predStepsLeft[$\gamma$] $\times$ req.numTokens
\EndFunction

\Function{ClusterCapacity}{\;}
    \State totalCapacity $\gets 0$
    \For{each instance}
        \State tp $\gets$ instance.tpDegree
        \State capacity $\gets$ \smash{\ensuremath{
            \max_{\text{batch}}
            \left(\frac{\text{batch}}{\text{latencyProfile}[\text{tp}][\text{batch}] }\right)
            \times \text{SLO}
        }}
        \State totalCapacity $\gets$ totalCapacity + capacity
    \EndFor
    \State \Return totalCapacity
\EndFunction
\Function{UpdateAllowedConfThresholds}{\;}
    \State $R \gets$ all requests, $n \gets |R|$

    \State $\Gamma \gets$ default allowed confidence thresholds
    \State $\gamma_{\min} \gets$ lowest value in $\Gamma$

    \State minRemainLoad $\gets$ array of length $n$ initialized to $0$

    \For{$i = n-2$ \textbf{downto} $0$}
        \State minRemainLoad[$i$] $\gets$ minRemainLoad[$i{+}1$]
              $+$ \Call{ReqLoad}{$R[i{+}1]$, $\gamma_{\min}$}
    \EndFor

    \State usedLoad $\gets 0$, totalCapacity $\gets$ \Call{ClusterCapacity}{\;}

    \For{$i = 0$ \textbf{to} $n{-}1$}
        \State req $\gets R[i]$
        \State req.allowedConfThresh $\gets \Gamma$

        \For{$\gamma$ in desc $\Gamma$}
            \State reqLoad $\gets$ \Call{ReqLoad}{req, $\gamma$}
            \State remainLoad $\gets$ minRemainLoad[$i$]

            \If{usedLoad + reqLoad + remainLoad $\le$ totalCapacity
                        \textbf{or} $\gamma = \gamma_{\min}$}
                \State usedLoad $\gets$ usedLoad $+$ reqLoad
                \State \textbf{break}
            \Else 
                \State remove $\gamma$ from req.allowedConfThresh
            \EndIf
        \EndFor
    \EndFor
\EndFunction
\end{algorithmic}
\end{algorithm}
\revadd{Algorithm~\ref{alg:update_allowed_conf_thresh} outlines this adaptive load control mechanism. The algorithm iterates through all active requests in order of arrival time. For each request, it iterates through the candidate confidence thresholds in descending order. For each threshold, it computes the worst-case aggregate load as the sum of (1) the load already committed to previously processed requests, (2) the load of the current request if it were to use this threshold, and (3) a conservative estimate of the load from remaining requests, computed under the assumption that all remaining requests use the lowest confidence threshold. If this worst-case load exceeds cluster capacity, the algorithm removes this threshold from the request's allowed set and continues to the next lower threshold; otherwise, it keeps this threshold and moves on to the next request. By doing so, the algorithm finds the widest range of confidence thresholds for each request that keeps the aggregate load within cluster capacity, thereby maximizing quality for individual requests while ensuring overall system stability.}
\subsubsection{\texorpdfstring{\revadd{Step Prediction}}{Step Prediction}}
\label{subsubsec:tediserve/threshold_adjustment/step_prediction}
\begin{figure}[!t]
    \centering
    \includegraphics[width=0.92\linewidth]{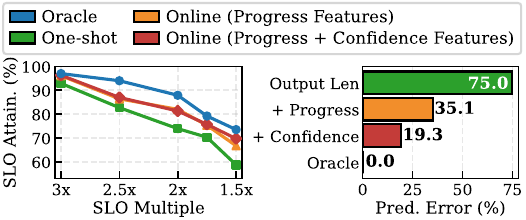}
    \caption{Serving performance of LLaDA on GSM8K (left) and step prediction error (right) under different step prediction strategies.}
    \label{fig:step_predictor_motivation}
\end{figure}

\revadd{Both SLO-aware confidence threshold adjustment (\S\ref{subsubsec:tediserve/threshold_adjustment/slo_aware_conf_adjustment}) and adaptive load control (\S\ref{subsubsec:tediserve/threshold_adjustment/load_control}) rely on estimates of how many denoising steps remain for a request under a candidate confidence threshold. A naive approach is to make a one-shot estimate when a request arrives using the request's output length and the historical average number of steps required under each confidence threshold. As shown in Figure~\ref{fig:step_predictor_motivation}, this one-shot baseline can be inaccurate, with prediction error of up to 75\%. During serving, such errors may cause the controller to select suboptimal confidence thresholds, reducing SLO attainment by up to 15 percentage points compared with an oracle that has exact knowledge of the remaining steps.}

\revadd{To improve estimation accuracy, we instead update the estimates online as the request progresses through the denoising process. Online updates allow the estimator to use runtime signals that are unavailable at request arrival, including the request's denoising progress and token-confidence features. Progress features (i.e. the fraction of tokens unmasked so far) capture how far a request has advanced in the denoising process: requests with a larger fraction of unmasked tokens are typically closer to completion. As shown in Figure~\ref{fig:step_predictor_motivation}, adding these features reduces prediction error by 34.9\%. Confidence-based features (i.e. the confidence values of the output tokens) further captures the uncertainty of the DLM. Intuitively, when the DLM is more certain about the predicted tokens, fewer denoising steps are typically needed to finalize the output. These features further reduce prediction error by 15.8\%. As shown in Figure~\ref{fig:step_predictor_motivation}, the resulting improvement in prediction accuracy translates into better serving performance, improving SLO attainment by up to 11 percentage points and generation accuracy by up to 1.1\% compared to one-shot estimation.}

We use these features to train a Step Predictor that estimates the remaining work of a request under each candidate confidence threshold. Formally, let $S_r(t,\gamma)$ denote the true number of remaining steps for request $r$ at denoising step $t$ under confidence threshold $\gamma$, and let $\mathbf{x}_r(t)$ denote the feature vector used by the predictor. This vector consists of the output length, denoising progress, and token-confidence features observed during denoising (the full list of features is provided in Section~\ref{sec:step-predictor-features}). We train a predictor $g$ that takes as input the request state $(t,\gamma,\mathbf{x}_r(t))$ and estimates the number of remaining denoising steps: $\hat{S}_r(t,\gamma) = g(t,\gamma,\mathbf{x}_r(t))$. We train $g$ offline as a gradient boosting machine~\cite{friedman2001greedy} using a small training set of 100 requests, where each request is evaluated under every candidate confidence threshold configured by the system administrator. This profiling and training process only needs to be performed once per deployed model.

\begin{figure}[t]
    \centering
    \includegraphics[width=0.48\linewidth]{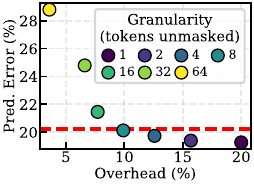}
    \caption{Step prediction error and execution overhead at different prediction granularities. Here, execution overhead is measured as the time spent generating predictions using the predictor relative to the execution time of a denoising step. The red dashed line marks a 5\% relative error upper bound over the 1-token-granularity.}
    \label{fig:step_predictor_granularity}
\end{figure}
\revadd{During deployment, however, updating step estimates after every denoising step (i.e., every time a token is unmasked) can be prohibitively expensive. As shown in Figure~\ref{fig:step_predictor_granularity}, per-token updates introduce an execution overhead of 20\%. Since this overhead lies on the execution hot path, it can significantly affect serving performance. To reduce this overhead, we observe that the predicted number of remaining steps does not change substantially after every token unmasking. As shown in Figure~\ref{fig:step_predictor_granularity}, updating the prediction only after several newly unmasked tokens substantially reduces scheduling overhead while incurring only a small increase in prediction error. In our setting, we select the largest update granularity (i.e., the number of newly unmasked tokens between updates) whose prediction error remains within 5\% of the per-token baseline. To further reduce overhead, we perform predictions in a separate asynchronous worker thread and batch them across multiple requests and all candidate confidence thresholds to amortize the cost.}
\subsubsection{\texorpdfstring{\revadd{Request Scheduling}}{Request Scheduling}}
\label{subsubsec:tediserve/threshold_adjustment/scheduling}
\begin{algorithm}[t]
\caption{Request Scheduling}
\label{alg:scheduling}
\setlength{\textfloatsep}{0ex}
\begin{algorithmic}[1]

\Function{FindBestInstance}{nTok}
    \State $I \gets \smash{\{\text{inst}\mid
	    \text{inst.batchSize}+\text{nTok}\le\text{inst.maxBatchSize}\}}$

    \If{$I = \emptyset$}
        \State \Return \textsc{None}
    \EndIf
    \State \Return the inst $\in I$ minimizing 
        latencyProfile[inst.tpDegree][inst.batchSize+\text{nTok}]
\EndFunction

\Function{Schedule}{\;}
    \State $R \gets$ unscheduledRequests
    \State $R \gets R \cup$ finishedStepRequests

    \ForAll{req in $R$}
        \State inst $\gets$ \Call{FindBestInstance}{req.nTok}
        \If{inst $\neq$ \textsc{None}}
            \State req.instance $\gets$ inst
        \EndIf
    \EndFor

    \ForAll{req in $R$ such that req.instance $\neq$ \textsc{None}}
        \If{req was migrated}
            \State \Call{UpdateConfThreshold}{req}
        \EndIf
    \EndFor
\EndFunction
\end{algorithmic}
\end{algorithm}
\revadd{Once the confidence thresholds have been determined, the Scheduler decides which model instance executes each request's next denoising step. Because the system maintains multiple model instances running asynchronously, scheduling decisions are triggered whenever a new request arrives or when any model instance completes a denoising step. Several such events may occur in close succession, so the scheduler may handle multiple requests during a single decision cycle.}

\revadd{The scheduling algorithm is outlined in Algorithm~\ref{alg:scheduling}. At each scheduling event, the Scheduler processes unscheduled requests -- either newly arrived requests or requests that were previously deferred because no model instance could accommodate them -- and requests that have just completed a denoising step. For each request, the Scheduler assigns it to the model instance with the lowest denoising-step latency, favoring instances with higher TP degrees or lighter load and balancing load across the cluster.}

\revadd{A request may be migrated between model instances between denoising steps. Because individual denoising steps are essentially stateless without KV caching, migration is extremely lightweight: the system only needs to send the prompt and the current masked/unmasked tokens to the new executor, a transfer size on the order of tens of kilobytes even for long sequences. This enables denoising-step-level migration and fine-grained load balancing, which we show in Section~\ref{sec:appendix/migration_granularity} can reduce tail latency by up to 19\%.}

\revadd{Whenever a request is assigned or migrated to a new model instance, its confidence threshold is updated to reflect the change in predicted remaining latency, which is affected by the new instance's TP degree and load.}
\subsection{Reconfiguration}
\label{subsec:tediserve/reconfiguration}
\begin{table}[tbp]
    \centering
    \addtolength{\tabcolsep}{-0.5em}
    \begin{tabular}{P{0.20\linewidth} p{0.79\linewidth}}
        \hline
        Symbol & Definition \\
        \hline
        $\mathcal{C}$ & Set of possible confidence thresholds \\
        $\mathcal{D}$ & Set of possible TP degrees \\
        $\mathcal{B}$ & Set of possible batch sizes \\
        $\mathcal{N}$ & Set of compute nodes \\
        $\mathcal{K}$ & Set of input classes \\
        $SLO$ & Latency SLO \\
        $P_k$ & Prompt length of class $k \in \mathcal{K}$ \\
        $O_k$ & Output length of class $k \in \mathcal{K}$ \\
        $RPS_k$ & Arrival rate of class $k \in \mathcal{K}$ \\
        $T_c$ & \makecell[l]{Average number of tokens unmasked per step\\under confidence threshold $c \in \mathcal{C}$} \\
        $L_{d,b}$ & Latency per step for TP-$d$ and batch size $b$ \\
        $|\mathcal{G}_n|$ & Number of GPUs on node $n \in \mathcal{N}$ \\
        \hline
        $s_{k,c} \in \mathbb{R}_{\geq 0}$ & Steps of class $k$ using confidence threshold $c$ \\
        $s_{k,d} \in \mathbb{R}_{\geq 0}$ & Steps of class $k$ using TP degree $d$ \\
        $t_k \in \mathbb{R}$ & Latency of class $k$\\
        $z_{d,b} \in \mathbb{B}$ & Whether TP-$d$ machines use batch size $b$ \\
        $x_{n,d} \in \mathbb{N}$ & Number of TP-$d$ instances on node $n$ \\
        \hline
    \end{tabular}
    \caption{Notation used in Stage 1 formulation.}
    \label{tab:stage1_notation}
    \vspace{-3ex}
\end{table}
To adjust the cluster configuration in response to changing workload, the Workload Monitor groups requests by their input-output sizes and tracks the arrival rate of each class. This information is then used by the Reconfigurator, which periodically (e.g., every 5 minutes, at a much coarser granularity than the Scheduler) computes an updated cluster configuration suited to the current workload. Determining this configuration is formulated as a two-stage ILP.\\
\textbf{Stage 1: Quality-Aware Capacity Planning} In the first stage, the goal is to determine the number of model instances at each TP degree such that all request SLOs are met and the cluster has sufficient capacity to sustain the workload in steady state (i.e., without unbounded queue growth). If the optimization problem is infeasible under these constraints, we fall back to using only TP-1 instances, which maximizes throughput, as discussed in \S\ref{subsubsec:challenges/cluster_level_serving}.

A trivial feasible solution is to force all requests to use the lowest confidence threshold, which minimizes load, but this severely reduces overall generation quality. Instead, the optimization seeks a configuration that maximizes the use of high confidence thresholds while still satisfying SLO and throughput constraints. Among all configurations achieving the same average confidence threshold, we further prefer those that minimize request latency. 

Table~\ref{tab:stage1_notation} summarizes the parameters and variables used in the formulation. Formally, the optimization problem has a hierarchical two-level objective. {\sloppy The primary objective is to maximize the average confidence threshold used per output token: $\max \sum_{k \in \mathcal{K}} RPS_k \cdot \frac{\sum_{c \in \mathcal{C}} c\, T_c\, s_{k,c}}{O_k}$. The secondary objective, applied only among solutions that tie on the first criterion, is to maximize SLO slack: $\max \sum_{k \in \mathcal{K}} RPS_k \cdot \left(SLO - t_k\right)$. The optimization problem is subject to the following constraints:}
{\sloppy
\begin{itemize}[leftmargin=0.1in]
    \item \textbf{Token-Unmasking Completeness:} For each request, the total number of tokens unmasked across all confidence thresholds must equal the output length:
    $\sum_{c \in \mathcal{C}} T_c\, s_{k,c} = O_k, \; \forall k \in \mathcal{K}.$
    \item \textbf{Step Allocation Consistency:} For each request, the total number of denoising steps allocated across confidence thresholds must match the total allocated across TP degrees:
    $\sum_{c \in \mathcal{C}} s_{k,c} = \sum_{d \in \mathcal{D}} s_{k,d}, \; \forall k \in \mathcal{K}.$
    \item \textbf{SLO Satisfaction:} For each request, the end-to-end latency of executing all denoising steps, under the selected TP degree and batch size, must not exceed its latency SLO:
    $t_k = \sum_{d \in \mathcal{D}} \sum_{b \in \mathcal{B}} z_{d,b}\, L_{d,b}\, s_{k,d} \le SLO_k, \; \forall k \in \mathcal{K}.$
    \item \textbf{Batch Size Selection}: All model instances of a given TP degree share a single batch size configuration. This is not a fundamental requirement of the system but a modeling simplification that keeps the optimization problem manageable:
    $\sum_{b \in \mathcal{B}} z_{d,b} = 1, \; \forall d \in \mathcal{D}.$
    \item \textbf{Node GPU Capacity}: For each node, the total GPU demand of all model instances placed on it must not exceed its GPU capacity:
    $\sum_{d \in \mathcal{D}} x_{n,d}\, d \le |\mathcal{G}_n|, \; \forall n \in \mathcal{N}.$
    \item \textbf{System Stability}: For each TP degree, the load assigned to it does not exceed its processing capacity. This ensures that the system remains stable with bounded queue growth, according to queueing theory~\cite{kleinrock1975queue}:
    $\sum_{k \in \mathcal{K}} RPS_k\, s_{k,d}\,(P_k + O_k) \le \sum_{b \in \mathcal{B}} z_{d,b}\, b\, \left( \frac{1}{L_{d,b}} \sum_{n \in \mathcal{N}} x_{n,d} \right), \; \forall d \in \mathcal{D}.$
\end{itemize}
}
\begin{table}[tbp]
    \centering
    \addtolength{\tabcolsep}{-0.4em}
    \begin{tabular}{P{0.22\linewidth} p{0.77\linewidth}}
        \hline
        Symbol & Definition \\
        \hline
        $\mathcal{I}$ & Set of old model instances \\
        $\mathcal{G}_n$ & Set of GPUs on node $n \in \mathcal{N}$ \\
        $\mathcal{L}_{n,d}$ & Set of slots for TP-$d$ on node $n$ ($\mathcal{L}_{n,d} = \{1, 2, ...,\lfloor |\mathcal{G}_n|/d \rfloor\}$) \\
        $\mathcal{G}_i$ & Set of GPUs originally used by old instance $i$ \\
        $(n_i, d_i)$ & Node and TP degree of model instance $i$ \\
        $R_d$ & Required number of TP-$d$ instances ($R_d = \sum_{n \in \mathcal{N}} x_{n,d}$) \\
        \hline
        $u_{n,d,\ell} \in \mathbb{B}$ & Whether slot $\ell$ for TP-$d$ on node $n$ is activated \\
        $a_{n,d,\ell,g} \in \mathbb{B}$ & Whether GPU $g \in \mathcal{G}_n$ is assigned to slot $(n,d,\ell)$ \\
        $m_{i,\ell} \in \mathbb{B}$ & Whether executor $i$ is preserved by mapping to slot $(n_i, d_i, \ell)$ \\
        \hline
    \end{tabular}
    \caption{\revshep{Notation used in Stage 2 formulation.}}
    \label{tab:stage2_notation}
    \vspace{-4ex}
\end{table}
\textbf{Stage 2: Minimal-Overhead Reconfiguration Planning} The Stage-1 solution specifies a cluster configuration through the variables $x_{n,d}$. Because all nodes are symmetric and therefore interchangeable in a homogeneous cluster, only the total number of TP-$d$ model instances matters. Let $R_d:= \sum_{n \in \mathcal{N}} x_{n,d}$ denote the required number of TP-$d$ instances. If the current configuration already matches $R_d$ for all $d \in \mathcal{D}$, no reconfiguration is needed. Otherwise, we must terminate and launch model instances to reach the target configuration, and we solve a second ILP to determine the minimal-overhead reconfiguration plan. 

The objective of Stage 2 is to maximize the number of model instances that can be preserved (i.e., instances that continue using the same GPUs on the same node) while satisfying $R_d$. Preserving more instances directly reduces the number of terminations and launches required during the transition, which in turn limits the temporary reduction in available processing capacity during reconfiguration and minimizes the impact on service availability.

\revshep{To express this problem cleanly, we introduce the notion of \emph{slots}. For each node $n$ and TP degree $d$, we define a set $\mathcal{L}_{n,d}$ of potential TP-$d$ slots, where each slot represents a feasible placement location for a TP-$d$ model instance. A node with $|\mathcal{G}_n|$ GPUs can host at most $\lfloor |\mathcal{G}_n| / d \rfloor$ such slots. These slots serve as fixed candidate containers to which GPUs may be assigned and old instances may be matched.}

\revshep{Table~\ref{tab:stage2_notation} summarizes the parameters and variables used in this formulation. The objective is to maximize the number of preserved model instances, $\max \sum_{i \in \mathcal{I}} \sum_{\ell \in \mathcal{L}_{n_i,d_i}} m_{i,\ell}$, subject to the following constraints:}
\vspace{-0.5ex}
{\sloppy
\begin{itemize}[leftmargin=0.1in]
    \item \revshep{\textbf{Stage 1 Compliance:} For each TP degree $d$, the number of slots activated for TP-$d$ is equal to $R_d$, the plan generated by Stage 1:
    $\sum_{n \in \mathcal{N}} \sum_{\ell \in \mathcal{L}_{n,d}} u_{n,d,\ell} = R_d, \; \forall d \in \mathcal{D}.$}
    \item \revshep{\textbf{Slot GPU Assignment:} If a slot of TP-$d$ is activated on a node, it must be assigned exactly $d$ GPUs from that node:
    $\sum_{g \in \mathcal{G}_n} a_{n,d,\ell,g} = d \cdot u_{n,d,\ell}, \; \forall n \in \mathcal{N},\, \forall d \in \mathcal{D},\, \forall \ell \in \mathcal{L}_{n,d}.$}
    \item \revshep{\textbf{GPU Exclusivity:} Each GPU on a node can only be assigned to one slot:
    $\sum_{d \in \mathcal{D}} \sum_{\ell \in \mathcal{L}_{n,d}} a_{n,d,\ell,g} = 1, \; \forall n \in \mathcal{N},\, \forall g \in \mathcal{G}_n.$}
    \item \revshep{\textbf{Preservation Uniqueness:} If a model instance is preserved, it can only be assigned to one slot:
    $\sum_{\ell \in \mathcal{L}_{n_i,d_i}} m_{i,\ell} \le 1, \; \forall i \in \mathcal{I}.$}
    \item \revshep{\textbf{Preservation Implies Slot Activation:} If a model instance is preserved, the corresponding slot needs to be activated:
    $m_{i,\ell} \le u_{n_i,d_i,\ell}, \; \forall i \in \mathcal{I},\, \forall \ell \in \mathcal{L}_{n_i,d_i}.$}
    \item \revshep{\textbf{Preservation Implies Same GPU Set:} If a model instance is preserved, all the GPUs it previously occupied must be exactly the GPUs assigned to its new slot:
    $m_{i,\ell} \le a_{n_i,d_i,\ell,g}, \; \forall i \in \mathcal{I},\, \forall \ell \in \mathcal{L}_{n_i,d_i},\, \forall g \in \mathcal{G}_i.$}
\end{itemize}
}
\vspace{-0.5ex}
\revshep{From the variables $u_{n,d,\ell}$ and $a_{n,d,\ell,g}$, we reconstruct the final cluster configuration by identifying which TP-$d$ slots are activated on each node and which GPUs are assigned to them. For any activated slot where $m_{i,\ell} = 1$, the corresponding old model instance $i$ is preserved and continues to use its original GPUs. For all remaining activated slots, a new model instance must be launched; consequently, any old instance currently occupying the GPUs assigned to that slot must be terminated.}\\
\textbf{Practical Consideration} To avoid frequent or unnecessary reconfiguration, the system triggers reconfiguration only when the Stage 1 proposal improves sufficiently over the current configuration (e.g., at least 10\% improvement in the primary or secondary objective). Furthermore, to ensure robustness against short-term load fluctuations, the request-class RPS values used in the optimization include a safety buffer, making them slightly higher than the observed RPS.

If a reconfiguration is deemed necessary, the system executes the plan generated by Stage 2 sequentially to prevent large temporary reductions in cluster processing capacity. New model instances are launched one at a time, and before each launch, only the minimum necessary set of old instances whose GPUs are required for that launch are terminated.
\subsection{Supporting Approximate KV Caching}
\label{subsec:tediserve/kv_caching}
With approximate KV caching, denoising steps split into two types: \emph{cache steps}, which reuse previously computed keys and values, and \emph{recompute steps}, which recompute the cache. Although TeDiServe’s core design remains effective, these two step types have substantially different computational costs and therefore require several adjustments to support approximate KV caching.\\
\textbf{Request Load, Cluster Capacity, and Predicted Time Left.}
Approximate KV caching makes denoising steps heterogeneous: cache steps compute only the tokens inside the current cache block, while recompute steps process all tokens in the sequence. Cache steps therefore incur a much lighter computational load and latency than recompute steps. To account for this heterogeneity, TeDiServe treats cache and recompute steps separately when estimating request load, cluster capacity, and remaining latency. For load control (\S\ref{subsubsec:tediserve/threshold_adjustment/load_control}, Algorithm~\ref{alg:update_allowed_conf_thresh}), the load of a request is computed as the combined load of its remaining cache and recompute steps, and cluster capacity is estimated using the corresponding block-level execution pattern. Similarly, when selecting the highest confidence threshold that still satisfies the SLO (\S\ref{subsubsec:tediserve/threshold_adjustment/slo_aware_conf_adjustment}, Algorithm~\ref{alg:selecting_conf_thresh}), TeDiServe estimates the remaining latency as the sum of the latency of the remaining cache and recompute steps. The detailed load, capacity, and remaining-time estimation algorithms are provided in Section~\ref{subsec:supplementary/kv_caching_algorithms}.\\
\textbf{Migration Granularity.} During request scheduling (\S\ref{subsubsec:tediserve/threshold_adjustment/scheduling}, Algorithm~\ref{alg:scheduling}), migrating a request during a cache step would require rebuilding its KV cache on the destination instance, effectively adding an additional recompute step. To avoid this overhead while still allowing migration for load balancing, we permit migration only at recompute steps. At these steps, the KV cache must be rebuilt regardless of migration, so moving the request incurs no additional cost. We show in Table~\ref{tab:eval/ablation/migration_granularity} of Section~\ref{sec:appendix/migration_granularity} that this reduces latency by up to 38\%.\\
\textbf{Reconfiguration.} In deriving the optimal cluster configuration (\S\ref{subsec:tediserve/reconfiguration}), instead of a single step variable $s$ and batch-size selector $z$, we use recompute steps $s$ and cache steps $s'$, together with batch-size choices $z$ and $z'$ for recompute and cache steps, respectively. Most constraints remain unchanged after substituting these variables. The primary modification is the System Stability constraint, which now accounts for the different computational costs of recompute and cache steps: {\sloppy $\sum_{k} RPS_k \Big[s_{k,d}(P_k + O_k)\,\alpha_d + s'_{k,d}\,B\,\beta_d\Big] \le \sum_{n} x_{n,d}, \; \forall d \in \mathcal{D}$.} Here, $\alpha_d = \sum_{b} z_{d,b}\,L_{d,b}/b$ is the per-step cost of recompute steps and $\beta_d  = \sum_{b} z'_{d,b}\,L_{d,b}/b$ is the per-step cost of cache steps,
and $B$ denotes the cache block size.
\vspace{-6pt}
\subsection{Extensions}
\label{subsec:tediserve/extension}
In this section, we discuss several extensions that TeDiServe can support with minimal modification.\\
\textbf{Non-homogeneous SLOs} Service providers may assign different latency SLOs to different requests -- for example, based on prompt/output length or service tier -- rather than using a single global SLO. Supporting heterogeneous SLOs requires only minor changes to TeDiServe. For capacity estimation (Algorithm~\ref{alg:update_allowed_conf_thresh}), one can use an average SLO across request classes, or a minimum SLO if prioritizing tighter-latency classes when estimating cluster capacity. In Stage~1 reconfiguration, each request class simply carries its own SLO. Importantly, the confidence-threshold selection and scheduling logic already operates on a per-request basis and uses each request’s remaining SLO budget. Thus, these components naturally extend to the non-homogeneous SLO setting without modification.\\
\textbf{Request Priority} Some deployments may prioritize certain requests, such as premium-tier users or quality-critical tasks, by ensuring they receive consistently higher generation quality. When determining the allowed confidence thresholds for each request (Algorithm \ref{alg:update_allowed_conf_thresh}), priority can be incorporated by ordering requests so that high-priority ones are processed first, giving them the widest feasible range of confidence thresholds. In the extreme, the system may guarantee that high-priority requests always retain access to the full range of confidence thresholds, while lower-priority requests adapt to keep the overall load within system capacity.\\ \revshep{Our evaluation assumes a single homogeneous latency SLO and equal priority across requests.}

\section{Implementation}
TeDiServe is built on top of vLLM, with approximately 9,500 lines of additional Python code. Model instances are launched as Ray~\cite{moritz2018ray} actors, and the Scheduler issues execution commands to them as Ray tasks. Each command contains the metadata required for a denoising step, including the request prompt and output length (if not already present), the set of tokens unmasked in the previous denoising step, and the selected confidence threshold.

With multiple model instances running concurrently, the Scheduler executes an asynchronous event loop, making scheduling decisions whenever new requests arrive or whenever any model instance completes a denoising step and returns control. The Workload Monitor continuously tracks workload statistics over a user-specified time window, and the Reconfigurator uses Gurobi~\cite{gurobi} as the underlying solver to compute an updated cluster configuration periodically.

To support approximate KV caching, FlashAttention~\cite{dao2022flashattention} is configured to use full bidirectional attention without causal masking. KV caches are stored in GPU memory and rebuilt during recompute steps as needed.

\section{Evaluation}
\label{sec:eval}
We evaluate TeDiServe on \revadd{16-GPU clusters} using a real-world LLM serving trace and diverse accuracy benchmarks. The evaluation shows that TeDiServe improves SLO attainment by up to 56.6 percentage points and reduces the end-to-end latency by \revadd{46\%}, with less than 1\% quality degradation compared to state-of-the-art serving systems that use a fixed confidence threshold.
\vspace{-6pt}
\subsection{Experimental Setup}
\label{subsec:eval/setup}
\begin{table}[t]
    \centering
    \begin{tabular}{l r r r r}
    \hline
    & \textbf{Avg} & \textbf{P50} & \textbf{P90} & \textbf{P99} \\
    \hline
    Prompt Length & 140 & 28 & 344 & 1698 \\
    Output Length & 264 & 224 & 544 & 928 \\
    \hline
    \end{tabular}
    \caption{ShareGPT statistics. Samples are filtered to exclude cases where prompt $+$ output exceeds the models' 4096-token limit. Output lengths are rounded up to the nearest multiple of 32 to match the denoising block size used for KV caching.}
    \label{tab:eval/setup/sharegpt}
    \vspace{-4ex}
\end{table}
\begin{table}[t]
    \centering
    \setlength{\tabcolsep}{3.5pt}
    \begin{tabular}{l l r r@{}}
    \hline
    \textbf{Benchmark} & \textbf{Domain} & \textbf{Samples} & \textbf{Prompt Len.} \\
    \hline
    GSM8K (5) ~\cite{cobbe2021gsm8k}     & Math     & 1,319  & 1,020 \\
    MMLU-Pro (0)~\cite{wang2024mmlupro}  & General  & 12,032  &   261 \\
    MBPP (3)~\cite{odena2021mbpp}     & Coding   & 474    &   737 \\
    \hline
    \end{tabular}
    \caption{Benchmarks used for evaluation. The number in parentheses denotes the number of few-shot examples.}
    \label{tab:eval/setup/benchmarks}
    \vspace{-5ex}
\end{table}
\textbf{Compute Platforms} \revadd{Experiments were conducted on two clusters. Experiments in \S\ref{subsec:eval/real_world_workload} and \S\ref{subsec:eval/ablation} were run on Cluster 1, while experiments in \S\ref{subsec:eval/accuracy_benchmarks} were run on Cluster 2. Cluster 1 consists of NVIDIA GH200 nodes. Each node contains four GH200 superchips, each equipped with one H100 GPU (96 GB HBM3) and one Grace CPU (72 cores), for a total of 4 GPUs, 288 CPU cores, 480 GB LPDDR5 system memory per node. Nodes are connected via a 200 Gbps interconnect. Cluster 2 consists of NVIDIA H100 nodes. Each node contains four H100 GPUs (96 GB HBM3), 96 vCPU cores, 1 TB DDR5 memory. Nodes are connected via a 100 Gbps interconnect.}\\
\textbf{Models} We evaluate our system on  LLaDA~\cite{nie2025llada} and Dream~\cite{ye2025dream}, two of the largest open-source DLMs with 8 billion and 7 billion parameters, respectively. Unless otherwise noted, we enable approximate KV caching with a denoising block size of 32 tokens and use a default confidence threshold of 0.9, following the configurations described in prior work~\cite{wu2025fastdllm}.\\
\textbf{Benchmark and Workloads} To evaluate TeDiServe in a realistic deployment setting, we use real-world LLM serving workloads derived from ShareGPT~\cite{sharegpt} and BurstGPT~\cite{wang2025burstgpt}. The ShareGPT dataset provides real conversational prompts and outputs but lacks arrival timestamps. In contrast, BurstGPT is a production LLM serving trace that contains request arrival timestamps but does not include prompts or outputs. We therefore combine the two: we use arrival timestamps from BurstGPT and sample prompt text and output lengths from ShareGPT (Table~\ref{tab:eval/setup/sharegpt}). BurstGPT contains two months of request arrivals. From this trace, we construct an ``average-day profile'' by averaging the requests-per-second (RPS) for each hour across the two months, which preserves the diurnal pattern of the workload. Following prior work~\cite{romero2021infaas}, we retain this shape but compress it into a 2-hour trace, where each 5-minute interval corresponds to one hour in the average-day profile. Within each 5-minute interval, arrivals are generated using a Poisson process with mean equal to the RPS of the corresponding hour. Finally, we scale the entire RPS profile \revshep{proportionally} (which peaks at \revadd{20}) to match the capacity of our cluster. \revshep{This scaling is uniform, so the trace retains the relative fluctuations of the original workload.} The final trace consists of \revadd{63,736} requests. 

For conversational workloads, there is no ground-truth answer for computing accuracy metrics. We therefore adopt an LLM-as-a-judge approach~\cite{zheng2023judgingllmasajudge, lin2023llmeval}. We use a strong LLM, \texttt{Llama-3.3-70B-Instruct}, as the judge model. For each prompt-response pair, it produces a quality score from 0 to 10. We report the average score across all requests as the generation-quality metric for a given serving system. The evaluation prompt for the judge is provided in Section~\ref{sec:llm-judge-prompt}. 

We also evaluate TeDiServe using three standard accuracy benchmarks spanning different domains, as shown in Table~\ref{tab:eval/setup/benchmarks}. For each benchmark, we use the same few shot settings as in prior work~\cite{nie2025llada, ye2025dream, wu2025fastdllm} and set the output length to 256. For each sample in the benchmarks, we send a corresponding request to the serving system. Although MBPP is a relatively large coding benchmark, its size is still insufficient for generating meaningful load at our cluster scale. To address this, we replicate each sample twice (three copies total) to create a request set large enough for load testing. Request arrivals follow a Poisson process~\cite{zhong2024distserve, sun2024llumnix} with varying request rates.

For each benchmark, by default we set the SLOs to $5\times$ the average latency of a single request running in isolation on a TP-1 model instance, similar to prior work~\cite{li2023alpaserve, stojkovic2025dynamollm, hong2025sola}. We also vary this SLO multiple to evaluate the system's ability to operate under stricter latency budgets.\\
\textbf{Baselines} To the best of our knowledge, TeDiServe is the first cluster-level serving system designed specifically for DLMs. \revproteus{We therefore adapt two representative serving systems, a general model serving system and an LLM serving system, to support DLMs and use them as our primary baselines. Both use a fixed confidence threshold. To isolate the benefit of TeDiServe's confidence threshold control, we additionally compare against a load-aware dynamic-threshold baseline inspired by Proteus~\cite{ahmad2024proteus} in our ablation study (\S\ref{subsec:eval/ablation}).}
\begin{itemize}[leftmargin=0.1in]
    \item \textbf{INFaaS}~\cite{romero2021infaas}: INFaaS is a general, multi-instance ML model serving system. Upon request arrival, INFaaS dispatches the request to the least-loaded worker for load balancing. We adapt it for DLM serving by using the number of tokens -- rather than the number of requests -- as the measure of load, which better reflects the computational demand of language model workloads.
    \item \textbf{Llumnix}~\cite{sun2024llumnix}: Llumnix is a multi-instance serving system for autoregressive language models. It uses efficient migration techniques to perform iteration-level load balancing while avoiding the eviction and preemption overheads caused by growing KV-cache memory. For DLMs, the output length (and thus the KV-cache size) is user-specified and known in advance, so this issue does not arise; however, Llumnix’s fine-grained load-balancing approach remains relevant. We adapt Llumnix for DLMs by restricting migration to only happen at recompute steps, which avoids the high overhead of recomputing KV-cache.
    \item \revproteus{\textbf{Proteus}~\cite{ahmad2024proteus}: Proteus is a model serving system that adapts to load through accuracy scaling: it periodically selects a mixture of model variants with different accuracy and throughput, maximizing overall accuracy while keeping cluster capacity above the incoming request rate. We adapt it to DLMs by treating confidence thresholds as the variants: using an offline profile of each threshold's accuracy and capacity, the baseline selects such a mixture of thresholds and assigns each arriving request a threshold from it that is fixed until completion.}
\end{itemize}
\begin{figure}[t]
    \centering
    \includegraphics[width=0.86\linewidth]{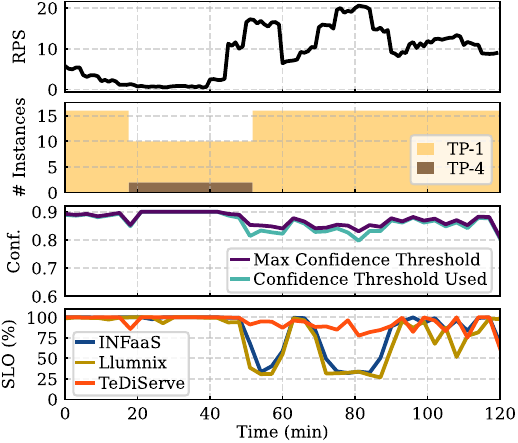}
    \vspace{-2ex}
    \caption{RPS of the trace, TeDiServe's cluster configuration and confidence threshold, and SLO attainment over time for all three systems on the real-world trace.}
    \label{fig:eval/sharegpt_timeline}
    \vspace{-2ex}
\end{figure}
\begin{table}[t]
    \centering
    \setlength{\tabcolsep}{3pt}
    \begin{tabular}{lcccccc}
        \hline
        \multirow{2}{*}{System} &
        SLO &
        \multirow{2}{*}{Score} &
        \multicolumn{3}{c}{Avg. Latency (s)} \\
        \cline{4-6}
         & Attain. & & Overall & Low-load & High-load \\
        \hline
        TeDiServe & 91.13\% & 6.80 & 5.75 & 2.26 & 6.33 \\
        Llumnix   & 60.98\% & 6.89 & 10.64 & 3.23 & 11.30 \\
        INFaaS    & 70.03\% & 6.89 & 8.80 & 2.71 & 9.34 \\
        \hline
    \end{tabular}
   \caption{Experimental results on the real-world trace. ``Score'' denotes the average evaluation score generated by the LLM judge. The low-load period corresponds to minutes 0 -- 45, and the high-load period corresponds to minutes 45 -- 120.}
    \label{tab:eval/sharegpt-results}
    \vspace{-4ex}
\end{table}
\begin{figure}[t]
    \centering
    \begin{subfigure}[b]{0.60\linewidth}
        \centering
        \includegraphics[width=\linewidth]{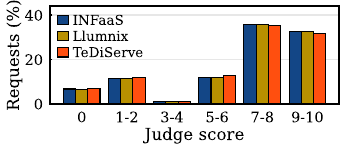}
        \caption{\revshep{Distribution of LLM judge scores}}
        \label{fig:eval/judge_score_distribution}
    \end{subfigure}
    \hfill
    \begin{subfigure}[b]{0.39\linewidth}
        \centering
        \includegraphics[width=\linewidth]{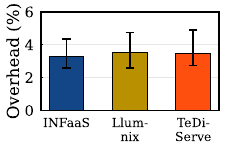}
        \caption{\revshep{Scheduling overhead}}
        \label{fig:eval/scheduling_overhead}
    \end{subfigure}
    \vspace{-5ex}
    \caption{\revshep{Quality and scheduling overhead on the real-world trace.}}
    \label{fig:eval/quality_and_latency}
    \vspace{-8pt}
\end{figure}
\begin{figure*}[h]
    \centering
    \includegraphics[width=0.91\textwidth]{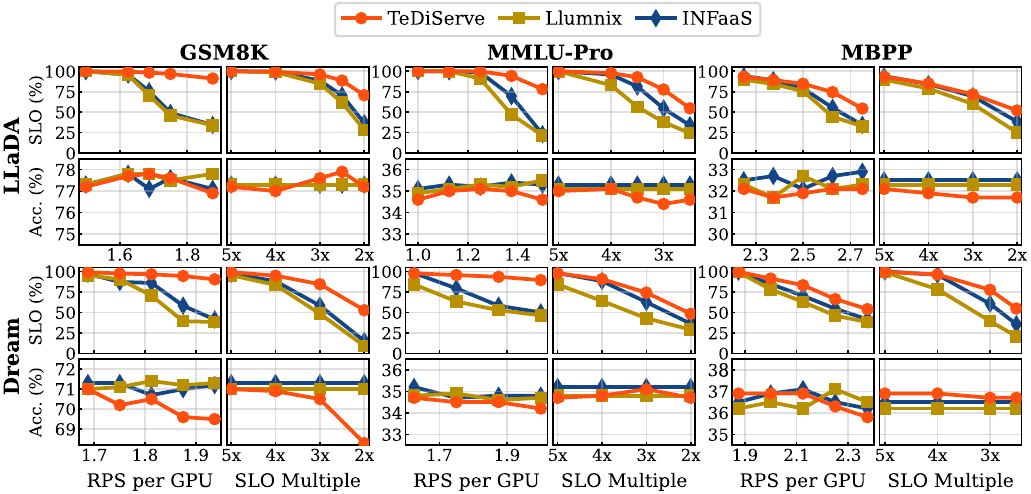}
    \caption{Experimental results for serving LLaDA (top) and Dream (bottom) on the three accuracy benchmarks.}
    \label{fig:eval/accuracy_benchmarks_combined}
\end{figure*}

\vspace{-6pt}
\subsection{Real-World Workload}
\label{subsec:eval/real_world_workload}
To validate TeDiServe’s effectiveness in a realistic deployment setting, we evaluate it by serving LLaDA using a trace constructed from the real-world ShareGPT and BurstGPT datasets \revadd{on a 16-GPU cluster}. As shown in the top figure of Figure~\ref{fig:eval/sharegpt_timeline}, the request load fluctuates over time, reflecting the diurnal pattern and the bursty nature of service usage. 
This allows us to evaluate the effectiveness of TeDiServe’s reconfiguration mechanism in responding to changing workload patterns. Since INFaaS and Llumnix do not support reconfiguration, we run both baselines with a fixed configuration of 16 TP-1 workers, which maximizes throughput and is best suited for handling peak loads. The generated response for each request is evaluated by the LLM judge on a scale of 0 to 10, with higher scores indicating better response quality.

The evaluation results are shown in Table~\ref{tab:eval/sharegpt-results}. TeDiServe achieves \revadd{21.1 percentage points} and \revadd{30.2 percentage points} higher SLO attainment and \revadd{34.7\%} and \revadd{46.0\%} lower latency, with only a \revadd{0.09-point drop in quality score} \revshep{and a similar score distribution (Figure~\ref{fig:eval/judge_score_distribution})} \revadd{compared to INFaaS and Llumnix, respectively.} Figure~\ref{fig:eval/sharegpt_timeline} shows that TeDiServe adapts both the cluster configuration and the maximum confidence threshold in response to fluctuating workload. At lower loads, requests can be served comfortably at the highest confidence threshold, and TeDiServe selects configurations with higher TP-degree model instances to reduce latency, yielding \revadd{16.6\%} and \revadd{30.0\%} lower latency compared to INFaaS and Llumnix, respectively. As the load increases, TeDiServe reconfigures to use more low TP-degree model instances for higher throughput, while lowering the maximum confidence threshold to keep effective load manageable. Under high load, more requests risk violating their latency SLOs, so they are assigned lower confidence thresholds to reduce the number of required denoising steps. \revadd{Together, these adaptations allow TeDiServe to maintain low latency and high SLO attainment \revshep{with negligible additional scheduling overhead (3.4\% vs.\ 3.3--3.5\% of each denoising step; Figure~\ref{fig:eval/scheduling_overhead})}, whereas the baselines suffer from queueing delays and increased load per model instance, leading to higher latency and a significant drop in SLO attainment. In the high-load period, TeDiServe achieves \revadd{32.2\%} and \revadd{44.0\%} lower latency compared to INFaaS and Llumnix, respectively.}
\vspace{-6pt}
\subsection{Accuracy Benchmarks}
\label{subsec:eval/accuracy_benchmarks}
To further evaluate TeDiServe's ability to maintain high SLO attainment while preserving accuracy, we evaluate it by serving LLaDA and Dream on three accuracy benchmarks spanning different application domains -- GSM8K~\cite{cobbe2021gsm8k} for math, MMLU-Pro~\cite{wang2024mmlupro} for general understanding, and MBPP~\cite{odena2021mbpp} for coding. To ensure a fair comparison with the baselines, all three systems are run with a fixed cluster configuration of 16 TP-1 workers and no reconfiguration, allowing us to assess their scheduling effectiveness under high request load.

Figure~\ref{fig:eval/accuracy_benchmarks_combined} shows the experimental results. As the request rate increases, request latency grows due to longer queueing delays and increased load per model instance, causing more requests to violate their SLOs. By dynamically adjusting the confidence threshold, TeDiServe effectively balances the trade-off between speed and quality, leading to consistently better serving performance across request rate and benchmarks. Under high request rates, TeDiServe achieves up to 55 percentage points higher SLO attainment with only a 0.7\% accuracy drop compared to INFaaS, and 56.6 percentage points higher SLO attainment with only a 0.9\% accuracy drop compared to Llumnix.

We also vary the SLO to evaluate each system’s ability to maintain performance under a fixed RPS but tighter latency requirements. TeDiServe consistently achieves higher SLO attainment as the SLO becomes stricter. In the moderately tight SLO regime ($\geq 3\times$), TeDiServe achieves up to 33.9 percentage points higher SLO attainment with only a 0.1\% accuracy drop compared to INFaaS, and 42.3 percentage points higher SLO attainment with only a 0.1\% accuracy drop compared to Llumnix. Under extremely tight SLOs, TeDiServe can further trade off quality for latency, allowing it to maintain higher SLO attainment.
\vspace{-6pt}
\subsection{Ablation Studies}
\label{subsec:eval/ablation}
\begin{figure}[t]    \centering
    \includegraphics[width=\linewidth]{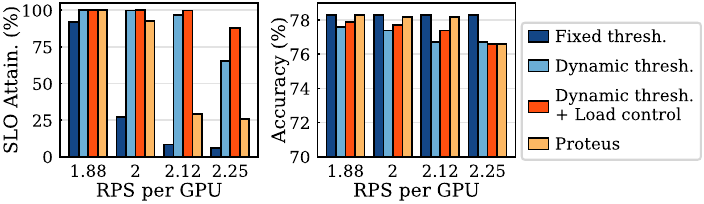}
    \caption{\revproteus{Confidence threshold control ablation serving LLaDA with GSM8K.}}
    \label{fig:eval/ablation/conf_adjustment}
    \vspace{-8pt}
\end{figure}
\revadd{\textbf{Confidence Threshold Control.} Figure~\ref{fig:eval/ablation/conf_adjustment} shows an ablation of TeDiServe's confidence-threshold control serving GSM8K requests with LLaDA. Compared to the fixed-threshold baseline, enabling SLO-aware dynamic threshold adjustment (\S\ref{subsubsec:tediserve/threshold_adjustment/slo_aware_conf_adjustment}) improves SLO attainment by up to 88.1 percentage points, while incurring at most a 1.6\% drop in accuracy. Incorporating aggregate-load-aware system-wide control (\S\ref{subsubsec:tediserve/threshold_adjustment/load_control}) further improves SLO attainment by up to 22.4 percentage points over dynamic threshold adjustment alone, while changing accuracy by at most 0.7\%.} \revproteus{We further compare the complete design against the Proteus-inspired baseline (\S\ref{subsec:eval/setup}). This single-level control improves SLO attainment by up to 65.6 percentage points over the fixed-threshold baseline, but TeDiServe attains up to 70.9 percentage points more than it, with at most a 0.8\% accuracy drop. Because the baseline fixes each request's threshold at arrival based only on the aggregate request rate, it cannot identify individual requests that fall behind their deadlines (e.g., due to queueing delays) or lower their thresholds during generation. Together, these results show that both per-request SLO-aware adjustment and system-wide load control are necessary for high SLO attainment.}\\
\begin{figure}[t]
    \centering
    \includegraphics[width=0.95\linewidth]{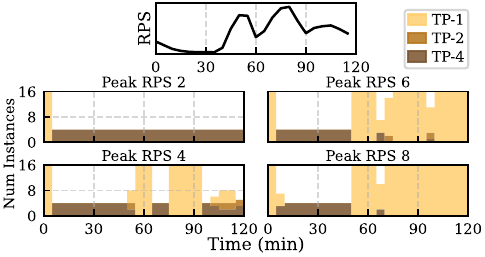}
    \vspace{-1ex}
    \caption{Derived cluster configuration over time under different peak RPS of the real-world workload.}
    \label{fig:eval/ablation/reconfig_ablation}
    \vspace{-2ex}
\end{figure}
\revadd{\textbf{Reconfiguration.} Figure~\ref{fig:eval/ablation/reconfig_ablation} evaluates TeDiServe's reconfiguration scheme under different load levels. In \S\ref{subsec:eval/real_world_workload}, we scale the real-world trace so that its peak RPS matches the capacity of our 16-GPU cluster. Here, we vary that peak RPS and simulate feeding the resulting trace into the Workload Monitor, while the Reconfigurator periodically uses the collected workload statistics to compute a new cluster configuration. The results show that the derived configurations adapt to both the overall load level and the temporal fluctuation in RPS. At low load, the reconfigurator favors fewer high-TP instances, which provide lower per-request latency. As the peak RPS increases, it shifts toward more low-TP instances to increase throughput and maintain SLO satisfaction under heavier demand. Across all settings, the chosen configuration also changes over time to track the changing request rate in the underlying trace.}\\
\begin{figure}[t]
    \centering
    \includegraphics[width=0.95\linewidth]{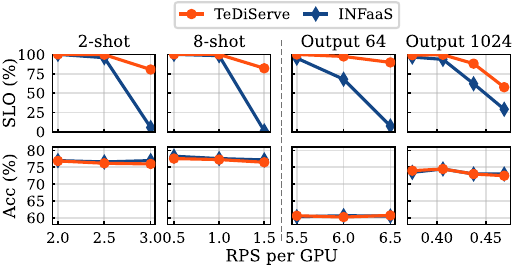}
    \vspace{-2ex}
    \caption{Input and output length ablation. For the input length ablation (left two columns), output length is fixed to 256. For the output length ablation (right two columns), number of few-shot examples is fixed to 5.}
     \label{fig:eval/ablation/input_output}
    \vspace{-2ex}
\end{figure}
\revadd{\textbf{Input and Output Length.} Figure~\ref{fig:eval/ablation/input_output} shows TeDiServe serving GSM8K requests with LLaDA under varying  input lengths (number of few-shot examples) and output lengths. Across all settings, TeDiServe consistently outperforms the fixed-threshold baseline (INFaaS) in SLO attainment regardless of input or output length. For varying input length, it improves SLO attainment by up to 81.0 percentage points with at most a 1.0\% accuracy drop. For varying output length, TeDiServe improves SLO attainment by up to 82.3 percentage points while incurring at most a 0.8\% accuracy drop. These results show that TeDiServe remains effective across a wide range of request sizes.}

\section{Related Work}
\textbf{Model Serving} There has been a long line of work for ML model serving, from general DNN serving~\cite{gujarati2020clockwork, shen2019nexus, romero2021infaas, crankshaw2017clipper, guo2022sommelier, ahmad2024proteus, zhang2023shepherd, li2023alpaserve}, to more specialized systems tailored for LLM~\cite{zhong2024distserve, stojkovic2025dynamollm, sun2024llumnix, kwon2023vllm, strati24dejavu, hong2025sola, patel2024splitwise, zhu2025nanoflow} and diffusion image models~\cite{ahmad2025diffserve, agarwal2024approx, li2025katz}. TeDiServe draws inspiration from these works on exploiting execution predictability~\cite{gujarati2020clockwork, zhang2023shepherd}, request scheduling~\cite{sun2024llumnix, romero2021infaas}, dynamic cluster reconfiguration~\cite{chang2025eva, mohan2022synergy} to design a high throughput language model serving system tailored for the unique characteristics of DLMs.\\
\textbf{Speed-Quality Tradeoff} The speed-quality trade-off is widely exploited across systems, including index search~\cite{mohoney2025quake}, video streaming~\cite{ganjam2015c3}, and online advertising~\cite{agarwal2014laser}. In model serving, systems such as Apparate~\cite{dai2024apparate} and $E^3$~\cite{padmanabha2024eednn} focus on this trade-off with early-exit DNN models, while Clipper~\cite{crankshaw2017clipper}, Proteus~\cite{ahmad2024proteus}, and Sommelier~\cite{guo2022sommelier} select among models to balance accuracy and efficiency. TeDiServe adopts a similar principle but departs from these prior works by controlling the speed-quality trade-off within a single model's execution, enabling high throughput and improved SLO attainment. \revshep{Proteus~\cite{ahmad2024proteus}, DiffServe~\cite{ahmad2025diffserve}, and MoDM~\cite{xia2026modm} share this broad goal but operate through model-variant selection: Proteus routes requests among variants with different speed-accuracy profiles, while DiffServe and MoDM combine lightweight and heavyweight image-diffusion models. These techniques are composable with TeDiServe, but differ in granularity. TeDiServe adjusts the confidence threshold within a single model execution, controlling both a request's remaining work and the aggregate cluster load without statically partitioning cluster capacity among variants.}\\
\revshep{\textbf{Diffusion Inference Optimizations} Recent work accelerates diffusion generation through caching and other inference optimizations, including Fast-dLLM~\cite{wu2025fastdllm}, dLLM-Cache~\cite{liu2025dllmcache}, dKV-Cache~\cite{ma2025dkvcache}, and Learning-to-Cache~\cite{ma2024learningtocache}. These optimizations are complementary to TeDiServe's cluster-level algorithms, which schedule and reconfigure across model instances. TeDiServe in fact builds on Fast-dLLM's approximate KV caching and explicitly models the cache-hit and recomputation heterogeneity it introduces.}

\section{Conclusion}
\label{sec:conclusion}
We presented TeDiServe, a cluster-level serving system for diffusion language models that achieves high throughput while meeting strict latency SLOs. TeDiServe dynamically adjusts the confidence threshold used during denoising for both aggregate load control and deadline-aware adaptation. It further adapts the cluster configuration to changing workloads and supports approximate KV caching. Our evaluation shows that TeDiServe improves SLO attainment by up to 56.6 percentage points with only 0.9\% quality degradation across diverse benchmarks and real-world workload.

\vspace{-6pt}
\section*{Acknowledgements}
We thank our anonymous shepherd and reviewers for their invaluable feedback. This work was supported by NSF Award CNS-2237306. We thank Zhao Zhang from Rutgers University for providing access to computing resources. This work used DeltaAI at the National Center for Supercomputing Applications (NCSA) through allocation CIS251202 from the Advanced Cyberinfrastructure Coordination Ecosystem: Services \& Support (ACCESS) program~\cite{boerner2023access}, which is supported by U.S. National Science Foundation grants \#2138259, \#2138286, \#2138307, \#2137603, and \#2138296.

\bibliographystyle{ACM-Reference-Format}
\bibliography{references}

\clearpage
\appendix
\section{Step Predictor Features}
\label{sec:step-predictor-features}
The Step Predictor uses 11 scalar features:
\begin{itemize}[leftmargin=0.1in]
    \item \nolinkurl{confidence_threshold}: the confidence threshold for the next denoising step.

    \item \nolinkurl{output_length}: the request's output length.

    \item \textbf{Progress features:}
    \begin{itemize}[leftmargin=0.1in]
        \item \nolinkurl{full_progress}: the completed denoising steps divided by the output length.
        \item \nolinkurl{full_unmask_progress}: the number of unmasked tokens divided by the output length.
        \item \nolinkurl{block_progress}: the current block index divided by the total number of blocks.
        \item \nolinkurl{block_unmask_progress}: the number of unmasked tokens in the current block divided by the block length.
    \end{itemize}

    \item \textbf{Confidence features:}
    \begin{itemize}[leftmargin=0.1in]
        \item \nolinkurl{output_min_confidence}: minimum confidence among all output tokens.
        \item \nolinkurl{output_q25_confidence}: lower-quartile token confidence.
        \item \nolinkurl{output_median_confidence}: median confidence among all output tokens.
        \item \nolinkurl{output_avg_confidence}: average confidence among all output tokens.
        \item \nolinkurl{output_q75_confidence}: upper-quartile token confidence.
    \end{itemize}
\end{itemize}

\section{Algorithm for Supporting Approximate KV Caching}
\label{subsec:supplementary/kv_caching_algorithms}
This section provides the detailed procedures to support approximate KV caching.\\
\textbf{Request Load and Cluster Capacity Calculation.}
\sloppy In the no-cache setting, throughput is approximated by $\max_{\text{batch}} \left(\frac{\text{batch}}{\text{latencyProfile}[\text{tp}][\text{batch}]}\right)$. With KV caching, however, this substantially overestimates throughput, because most steps are lightweight cache steps rather than compute-heavy recomputations. To address this, we estimate throughput at the \emph{block level}, assuming that each block consists of $(\text{cacheBlockSize} - 1)$ cache steps and one recompute step. The load of a cache step accounts for batching effects via the maximum number of sequences per batch, while the load of a recompute step corresponds to the average request length, obtained empirically. This block-based throughput estimate more accurately reflects the actual execution pattern under KV caching and provides a more reliable basis for capacity planning. The changes to load and capacity calculation are formalized in Algorithm~\ref{alg:load_capacity_kvcache}.
\begin{algorithm}[!t]
\caption{\small{Load and Capacity Calculation with KV Caching}}
\label{alg:load_capacity_kvcache}
\textit{Note:} LP is short for LatencyProfile. cacheBlockSize and maxNumSeqPerBatch are fixed system parameters.
\begin{algorithmic}[1]

\Function{ReqLoad}{req, $\gamma$}
    \State totalSteps $\gets$ req.predStepsLeft[$\gamma$]
    \State cacheSteps $\gets$ totalSteps - req.recompStepsLeft
    \State \Return
        cacheSteps $\times$ cacheBlockSize
        \;+\; req.recompStepsLeft $\times$ req.numTokens
\EndFunction

\Function{ClusterCapacity}{numTokenPerReq}
    \State totalCapacity $\gets 0$
    \For{each instance}
        \State tp $\gets$ instance.tpDegree
        \State cacheStepLoad $\gets$ cacheBlockSize $\times$ maxNumSeqPerBatch
        \State cacheStepLat $\gets$ LP[tp][cacheStepLoad]
        \State recompStepLat $\gets$ LP[tp][numTokenPerReq]
        \State $\mathrm{tput} \gets
            \frac{
                \text{cacheStepLoad} \times (\text{cacheBlockSize-1}) + \text{numTokenPerReq}
            }{
               \text{cacheStepLat} \times (\text{cacheBlockSize-1}) + \text{recompStepLat}
            }$
        \State capacity $\gets$ $\text{tput} \times \text{SLO}$
        \State totalCapacity $\gets$ totalCapacity + capacity
    \EndFor

    \State \Return totalCapacity
\EndFunction
\end{algorithmic}
\end{algorithm}\\
\textbf{Predicted Time Left.} Selecting the highest confidence threshold that still satisfies the SLO also depends on distinguishing cache and recompute steps. With KV caching, the remaining latency is the sum of the latency of all remaining cache steps and recompute steps. We estimate this conservatively by assuming that cache steps across requests are batched together and recompute steps are batched together, yielding worst-case latencies. This is formalized in Algorithm~\ref{alg:predict_time_left_cache}.

\begin{algorithm}[t]
\caption{Remaining Time Estimation with KV Caching}
\label{alg:predict_time_left_cache}
\begin{algorithmic}[1]

\Function{PredictedTimeLeft}{req, $\gamma$}
    \State inst $\gets$ req.instance
    \State tp    $\gets$ inst.tpDegree
    \State steps $\gets$ req.predStepsLeft[$\gamma$]
    \State cacheStepsLeft $\gets$ steps - req.recompStepsLeft
    \State $\smash{\text{cacheBatchSize}\gets \text{cacheBlockSize} \times \text{inst.numReqs}\,}$
    \State recompBatchSize $\gets$ inst.numTokens
    \State cacheStepLat  $\gets$ LP[tp][cacheBatchSize]
    \State recompStepLat $\gets$ LP[tp][recompBatchSize]
    \State \Return cacheStepsLeft $\times$ cacheStepLat + req.recompStepsLeft $\times$ recompStepLat
\EndFunction
\end{algorithmic}
\end{algorithm}

\section{LLM Judge Prompt}
\label{sec:llm-judge-prompt}

We use the following prompt for the LLM-based judge that scores assistant responses:

\begin{lstlisting}
You are grading an assistant response to a user request.

### USER REQUEST:
{prompt}

### ASSISTANT ANSWER:
{response}

---

Assign one integer score from 0 to 10 using this rubric:

- 10: Fully correct, directly answers the request, complete, clear, and safe.
- 7-9: Mostly correct and useful, with minor omissions or small inaccuracies.
- 4-6: Partially correct or helpful, but missing important content or containing notable mistakes.
- 1-3: Mostly incorrect, off-topic, or seriously incomplete.
- 0: Dangerous, nonsensical, or fails to answer the request.

Evaluate only these criteria:
1. Correctness
2. Relevance
3. Completeness
4. Clarity
5. Safety

Rules:
- Judge only the assistant answer.
- Prefer factual accuracy over style.
- Penalize unsafe or harmful advice heavily.
- If the request does not provide enough information to fully verify facts, score based on likely usefulness and internal consistency.
- Return only valid JSON matching the required schema.
- The score must be an integer from 0 to 10.
- The explanation must be one short sentence.

Return JSON only.

{{"score": <0-10>, "explanation": "<one short sentence>"}}.
\end{lstlisting}

\section{Experimental Results: Migration Granularity}
\label{sec:appendix/migration_granularity}
\begin{figure}[t]
    \centering
    \includegraphics[width=\linewidth]{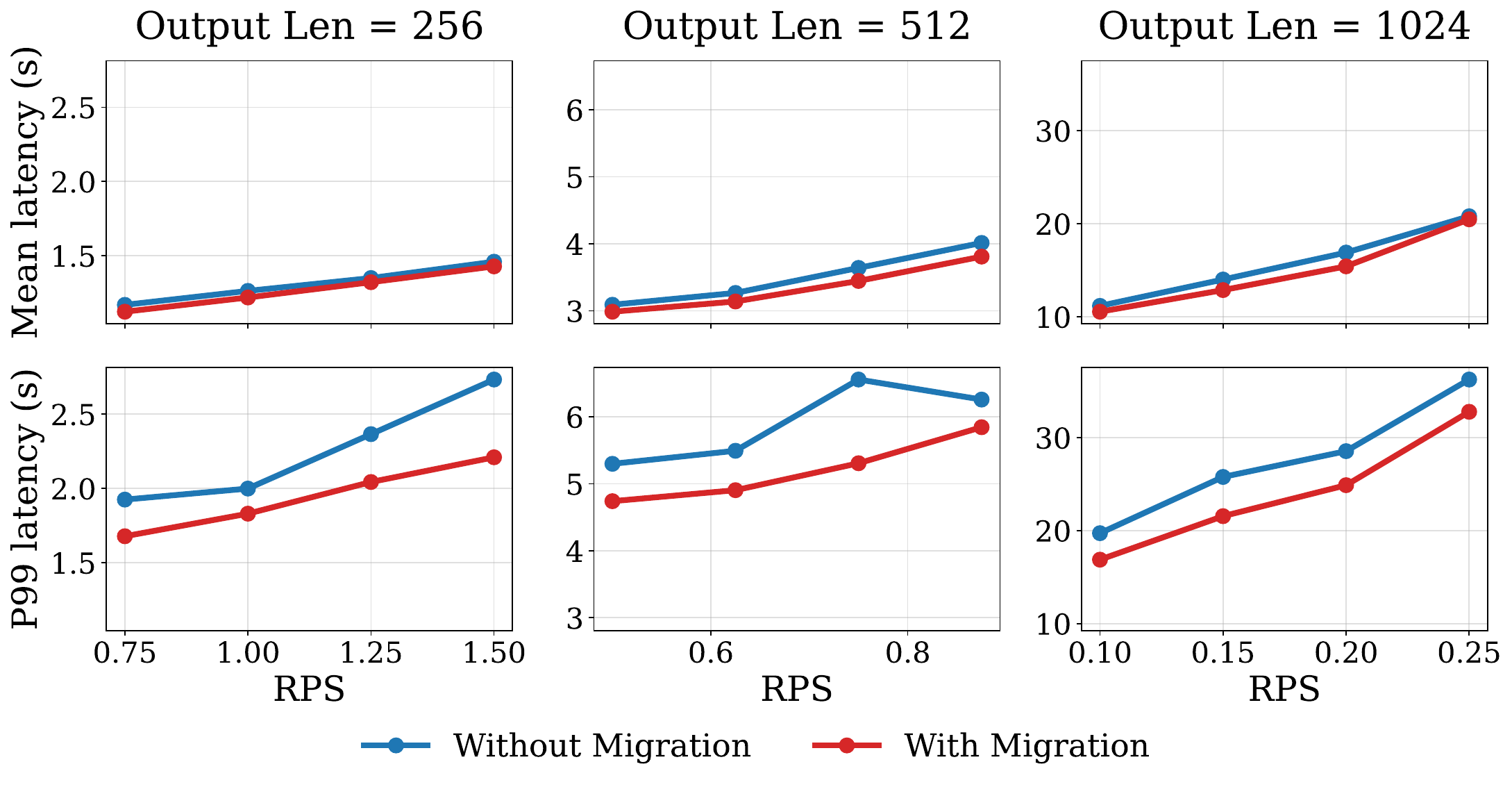}
    \caption{Migration effectiveness. Enabling denoising-step-level migration reduces P99 latency. Input length is set to 256 for all cases.}
    \label{fig:eval/ablation/migration_effectiveness}
\end{figure}
\begin{table}[t]
    \centering
    \renewcommand{\arraystretch}{0.85}
    \begin{tabular}{c c c c c}
        \toprule  %
        \# Model Instances & 4 & 8 & 12 & 16 \\
        \midrule  %
        \makecell[l]{Normalized latency \\ (vs.\ recompute-only)} 
            & 1.17 & 1.28 & 1.33 & 1.38 \\
        \bottomrule
    \end{tabular}
    \caption{Migration granularity. Shown is the request latency when migration is allowed at any step, normalized to TeDiServe's design where it is permitted only at recompute steps.}
    \label{tab:eval/ablation/migration_granularity}
\end{table}
TeDiServe performs migration at the denoising-step level to enable fine-grained load balancing beyond the initial dispatch. Figure~\ref{fig:eval/ablation/migration_effectiveness} shows that enabling migration reduces tail latency by up to 19\% across different RPS values and output lengths, as requests can opportunistically move to model instances that become idle or lightly loaded as other requests complete.
However, naively performing migration introduces non-negligible overhead when approximate KV caching is enabled, since migrating at non-recompute steps requires the KV cache to be rebuilt. As shown in Table~\ref{tab:eval/ablation/migration_granularity}, allowing migration at arbitrary steps increases request latency by up to 17\% with 4 model instances. This overhead grows with cluster size as more instances create more migration opportunities, reaching 38\% with 16 instances. As a result, TeDiServe only migrates a request at recompute steps, avoiding migration overhead while preserving the benefits of fine-grained load balancing.

\section{Experimental Results: Serving without Caching} 
\begin{figure}[t]
    \centering
    \includegraphics[width=0.73\linewidth]{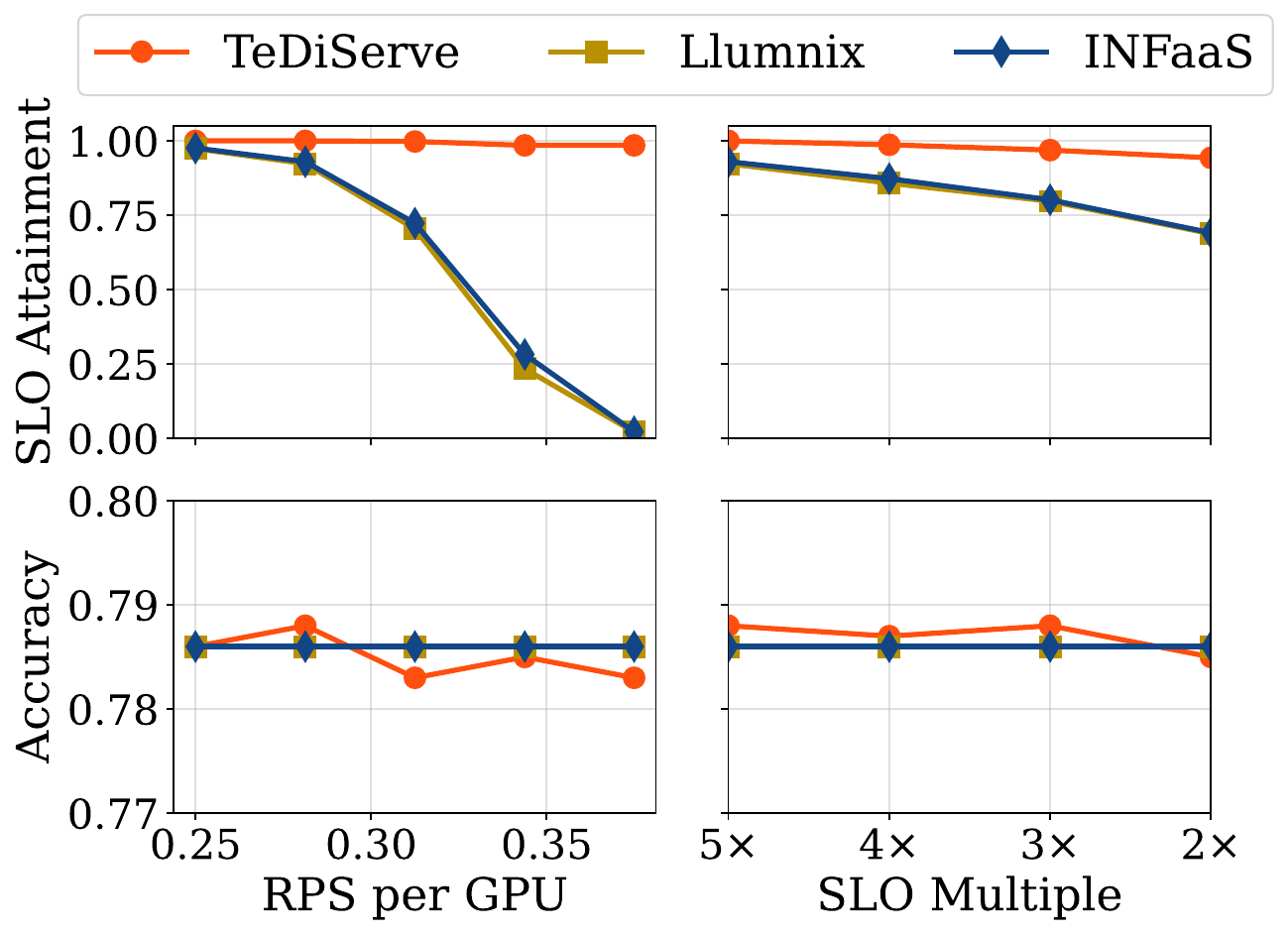}
    \caption{GSM8K without KV Caching.}
    \label{fig:eval/ablation/gsm8k_no_cache}
\end{figure}
Although TeDiServe includes support for approximate KV caching, its core design remains applicable regardless of whether caching is enabled. Figure~\ref{fig:eval/ablation/gsm8k_no_cache} shows the results of serving LLaDA without KV caching. In this setting, each denoising step becomes more computationally expensive; as a result, without adjusting confidence thresholds for load control, the queue grows quickly once the request rate exceeds cluster capacity, causing Llumnix and INFaaS to experience significant drops in SLO attainment. In contrast, TeDiServe maintains high SLO attainment without sacrificing accuracy. Under tight SLOs, it continues to perform robustly, achieving 25.5\% higher SLO attainment than Llumnix and INFaaS with no accuracy drop.

\section{Artifact Evaluation}
\label{sec:artifact-evaluation}

The TeDiServe artifact is available at\newline
\url{https://github.com/uw-mad-dash/TeDiServe} and archived at
\url{https://doi.org/10.5281/zenodo.22919779}.
It contains the scheduler,
step-time predictor, resource manager, and reconfiguration logic described in
Sections~3--4. The evaluator guide is in \texttt{artifact/README.md}, and the
evaluated version is tagged \texttt{eurosys27-ae}.\\
\textbf{Demonstration modes.}
The recommended \emph{smoke mode} runs the TeDiServe scheduler, control path,
step-time predictor, and measured latency-profile lookup with two simulated
model executors. No language model is loaded: the executors return fixed tokens
and confidence values. The smoke test therefore validates multi-instance
scheduling and request completion without requiring a GPU, CUDA, model weights,
Ray, or Gurobi. It does not evaluate model quality or real denoising behavior.

The optional \emph{GPU mode} runs actual LLaDA model inference, checks the
returned responses, and records per-instance work. We tested this mode on two
GH200 GPUs. Dynamic reconfiguration is disabled in both
modes, so Gurobi is needed only for separate reconfiguration experiments.\\
\textbf{Evaluator workflow.}
The public smoke image supports Linux x86\_64/AMD64 and ARM64/aarch64. The
primary Docker workflow is:
\begin{lstlisting}[language=bash]
docker pull ghcr.io/uw-mad-dash/tediserve/tediserve-smoke:2027
mkdir -p artifact-results
docker run --rm --user "$(id -u):$(id -g)" 
-v "$PWD/artifact-results:/opt/tediserve/artifact-results" 
ghcr.io/uw-mad-dash/tediserve/tediserve-smoke:2027
\end{lstlisting}

Once downloaded, the test takes under one minute and writes a machine-readable
result. A successful run completes all eight requests, assigns work to both
instances, and reports \texttt{passed: true}.

Evaluators with the GPU environment described in the README may additionally
run:
\begin{lstlisting}[language=bash]
bash artifact/run_demo.sh --mode gpu --instances 2 --requests 8 
--output-dir artifact-results/gpu
\end{lstlisting}
\textbf{Scope.}
The artifact exercises the central implementation path: deadline-aware request
scheduling across model instances using the learned step-cost predictor. It
also retains the measured LLaDA and Dream profiles used as scheduler inputs.

\end{document}